\documentclass[preprint,12pt]{elsarticle}

\usepackage{amssymb}
\usepackage{listings}
\usepackage{hyperref}
\usepackage{graphicx}
\usepackage{longtable}
\usepackage{tablefootnote}
\usepackage{tabularx}
\usepackage[T1]{fontenc}
\usepackage{booktabs}
\usepackage{lineno}

\journal{arXiv}

\begin{document}

\begin{frontmatter}



\title{XAI4Wind: A Multimodal Knowledge Graph Database for Explainable Decision Support in Operations \& Maintenance of Wind Turbines}


\author[1]{Joyjit Chatterjee*}
\ead{j.chatterjee-2018@hull.ac.uk}
% \ead[url]{www.cvr.cc, cvr@sayahna.org}
\author[2]{Nina Dethlefs}
\ead{n.dethlefs@hull.ac.uk}

\address{Department of Computer Science \& Technology, Dependable Intelligent Systems Research Group, University of Hull, Cottingham Road, Hull, HU6 7RX, United Kingdom}

% \ead[URL]{www.sayahna.org}

\begin{abstract}
% With the continuing rapid evolution of wind energy, there has been a significant rise in challenges pertaining to operations \& maintenance (O\&M) of wind turbines, which are complex engineering systems.
Condition-based monitoring (CBM) has been widely utilised in the wind industry for monitoring operational inconsistencies and failures in turbines, with techniques ranging from signal processing and vibration analysis to artificial intelligence (AI) models using Supervisory Control \& Acquisition (SCADA) data. However, existing studies do not present a concrete basis to facilitate explainable decision support in operations and maintenance (O\&M), particularly for automated decision support through recommendation of appropriate maintenance action reports corresponding to failures predicted by CBM techniques. Knowledge graph databases (KGs) model a collection of domain-specific information and have played an intrinsic role for real-world decision support in domains such as healthcare and finance, but have seen very limited attention in the wind industry. We propose XAI4Wind, a multimodal knowledge graph for explainable decision support in real-world operational turbines and demonstrate through experiments several use-cases of the proposed KG towards O\&M planning through interactive query and reasoning and providing novel insights using graph data science algorithms. The proposed KG combines multimodal knowledge like SCADA parameters and alarms with natural language maintenance actions, images etc. By integrating our KG with an Explainable AI model for anomaly prediction, we show that it can provide effective human-intelligible O\&M strategies for predicted operational inconsistencies in various turbine sub-components. This can help instil better trust and confidence in conventionally black-box AI models. We make our KG publicly available and envisage that it can serve as the building ground for providing autonomous decision support in the wind industry.\footnote{Cite as -- Chatterjee, J. and Dethlefs, N., “XAI4Wind: A Multimodal Knowledge Graph Database for Explainable Decision Support in Operations \& Maintenance of Wind Turbines”, \textit{arXiv e-prints}, 2020}    
\end{abstract}


% \begin{highlights}
% \end{highlights}


\end{frontmatter}

%\linenumbers

\section{Introduction}
In the global efforts towards tackling climate change and transitioning to sustainable energy sources, there has been a significant rise in developments and advancements of wind power systems in recent years \cite{darwish_al-dabbagh_2020}. However, the complexity of the environments in which wind turbines continue to be deployed, especially offshore regularly leads to irregular loads, operational inconsistencies and failures in the various electrical and mechanical sub-components within the turbine \cite{windenergy_journal}. Operations \& maintenance (O\&M) is integral to tackle such problems through fault detection and diagnosis \cite{ATTOUI201411}, and presently accounts for up to a third of the total cost of energy generation \cite{seyr_muskulus_2019}. 

Condition-based monitoring (CBM) \cite{STETCO2019620} plays a crucial role in O\&M, by facilitating monitoring of operational changes and anomalies in the turbine and its sub-components.
In the last decade, there has been a rising interest in utilising signal-processing techniques and physics-based modelling for CBM. More recently, the wind industry has witnessed application of Artificial intelligence (AI) techniques towards data-driven decision making in O\&M, by utilising the Supervisory Control \& Acquisition (SCADA) data generated by various sensors in the turbine at regular intervals. There have been some applications of Explainable AI (XAI) models \cite{BARREDOARRIETA202082,8329419,app10093258}, which can provide interpretations and reasoning behind predictions made by the conventionally black-box AI models during decision support, such as feature importances for SCADA parameters leading to faults \cite{windenergy_journal}. Such techniques have shown immense success, especially for anomaly prediction in turbine sub-components \cite{anomaly_paperhongshan,electronics9050751,Bo_2019}, wind resource assessment \cite{MURTHY20171320,DAI2017378} and forecasting vital operational parameters \cite{CHEN2018414,RePEc:eee:appene:v:241:y:2019:i:c:p:229-244,RePEc:eee:energy:v:201:y:2020:i:c:s0360544220308008} with high accuracy, which can be vital to improve the utilisation efficiency of wind energy \cite{LIU2020113324}. However, they fail to provide contextualised information (e.g. in the form of thorough descriptions of failures and maintenance reports to fix/avert such inconsistencies), which is essential to achieve ambient intelligence during decision support. More importantly, these techniques do not present a concrete basis to serve as the generic building ground for critical concepts in the wind industry, pertaining to maintenance strategies and their relationship with other sources of heterogeneous information, such as SCADA data from sensors, alarm logs, images describing failures and prognostic actions etc.

Knowledge bases (KBs) or knowledge graphs (KGs) play an integral role in supporting effective and informed decision making, by interlinking heterogeneous entities (such as numerical data from sensors) with contextualised and semantic information \cite{amador-paper-kg}. KGs have been successfully utilised for optimal decision support in several safety-critical applications such as healthcare for clinical decision support \cite{health_semantic}, power systems for optimising service processes \cite{Yuan_2018}, manufacturing for process safety management \cite{MAO2020107094}, finance for making business decisions \cite{kg_finance_yang} etc. Given the multidisciplinary nature of construction and management of wind farms as well as the presence of multiple sub-systems interacting with other sub-systems and various external components, a systematic organisation of knowledge in the wind industry can serve an integral role in assisting developers, designers and researchers to utilise such domain-specific information for decision support \cite{wes-5-259-2020}. 

In comparison to other disciplines, the wind industry has seen very limited application of KGs for explainable decision support. In an early work in this area, Zhu et al. \cite{4470371} developed a wind power plant information model by utilising the web ontology language (OWL) for systematic management of domain-specific information. The proposed ontology in the paper consists of conceptual information for wind power plant management with multiple classes based on different types of turbines, their functional parts etc., embedded as semantic information with corresponding OWL descriptions. While the paper demonstrates immense promise of graph databases for managing complex information in the wind industry through its conceptual overview, it does not provide a practical demonstration of the proposed ontology, including its use-cases in real-world information management in the wind energy domain. Moreover, there is no mention of the applicability of the proposed ontology for O\&M, in particular to facilitate decision support in real-world turbines.

Küçük and Arslan \cite{KUCUK2014484} utilised web resources, specifically from Wikipedia, to develop a domain ontology for the wind energy sector. Their ontology is constructed semi-automatically and provides a wide coverage of the wind energy domain as a discipline, including terms like meteorological data, sub-components of the turbine, monitoring and control system etc. This can be useful for utilisation in question-answering systems during information retrieval, wherein specific segments of natural language texts from the Wikipedia articles relevant to the end user can be accessed easily. While the semi-automatic construction of the ontology is less time consuming and a simpler process, it cannot provide low level details pertaining to the wind industry and O\&M of turbines. Moreover, it lacks the ability to provide a coherent view on critical aspects in CBM, including the role of SCADA data for anomaly prediction and power forecasting, maintenance tasks for failures in the turbine and its sub-components and more thorough and technically detailed descriptions for engineers \& technicians beyond the Wikipedia articles. The role of the KG for utilisation in real-world decision support systems is also not showcased in the paper. 

In another study, Quaeghebeur et al. \cite{wes-5-259-2020} proposed a graph database for the wind energy domain, consisting of a conceptual overview of several important terms and parameters in the wind energy sector, such as power curves, associated relationships between turbine sub-components, mathematical and statistical distributions for modelling vital parameters etc. The KG for this study was developed from the ground up by the authors, and provides a foundational ontology for the wind turbine as a system with a specific focus on offshore wind power. Similar to Zhu et al. \cite{4470371}, their focus is on developing knowledge representations and structured information pertaining to the wind industry. The paper provides significant information and descriptions which can be queried from the KG, being particularly beneficial for the purposes of education and training. However, the proposed KG does not contain any description of different types of faults which occur in real-world turbines, varieties of alarms and their relationship to turbine sub-components and maintenance action strategies to fix/avert the failures. More importantly, it does not demonstrate the applicability of the KG for O\&M, and its use-cases for explainable decision support in real-world practical scenarios.

As evident, while these few existing studies have demonstrated the promise of utilising KGs for systematically structuring conceptual information in the wind industry, they only serve as an endpoint \cite{amador-paper-kg} providing information for consultation without the ability to further reason over the data. Moreover, these ontologies need to be queried manually through specialised programming languages to extract relevant and meaningful information, which may not be easily accessible to turbine engineers \& technicians. The existing studies clearly do not focus on the critical aspects of O\&M, and their utility in decision-making systems for tasks such as recommendations and predictions. We believe that the optimal utility of KGs for decision support in the wind industry can only be realised by integrating conceptual information on O\&M with other heterogeneous data, such as SCADA parameters from sensors, alarm types, turbine sub-components preventive/predictive/corrective maintenance action strategies etc. and their associated relationships.

\begin{figure}[!h]
\centerline{\includegraphics[width=0.8\textwidth]{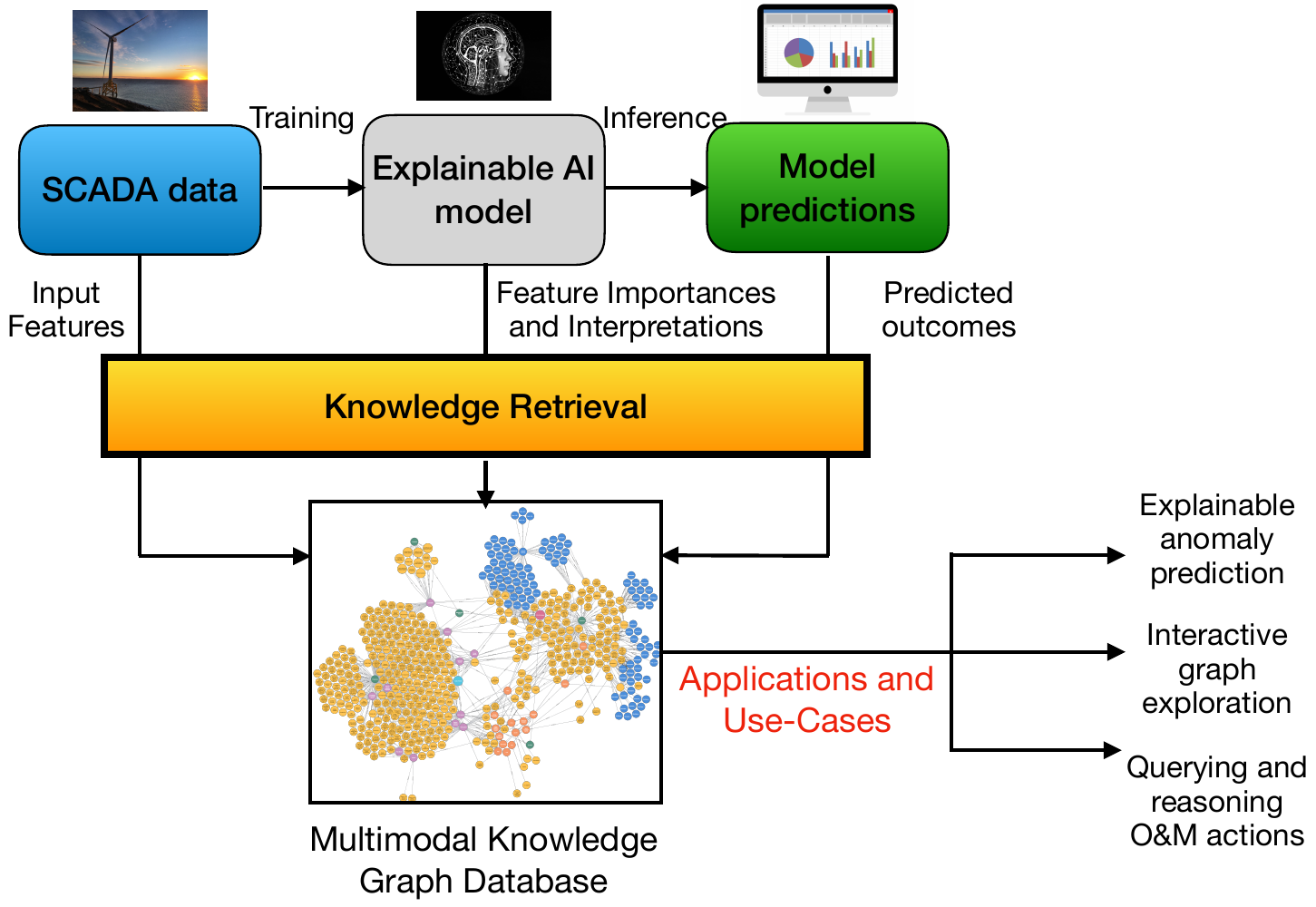}}
\caption{Overview of the proposed KG integrated with XAI models for explainable decision support.  \label{intro_kgfig}}
\end{figure}

In this paper, we propose the role of KGs for explainable decision support in the wind industry, and their utility in real-world applications for O\&M. We present a novel KG, \textit{XAI4Wind} and develop the domain-specific ontology for O\&M in the wind industry manually. The proposed KG presently contains 537 nodes and 1059 relationships (of 9 different types), and includes several types of heterogeneous data such as descriptions of various types of SCADA features, alarm types, turbine sub-components and maintenance action strategies. To demonstrate use-cases, we show that the KG can be queried interactively in Neo4j and graph data science algorithms can be applied to generate novel insights during O\&M. Notably, we utilise SCADA data from an operational turbine, and showcase its interfacing with an Explainable AI (XAI) model. We demonstrate the intrinsic role which the proposed KG can play in decision support during various failures predicted in the turbine by the XAI model, by providing thorough descriptions of fault conditions and maintenance actions required to fix/avert the failures. Unlike existing studies, our proposed KG can be queried both manually (by querying through relevant syntatic commands in programming) as well as automatically (through its integration with the XAI model), thereby paving the path to autonomous decision making for the wind energy sector which can serve as a complementary resource to aid decisions by engineers \& technicians. According to Futia and Vetrò \cite{info11020122}, integration of KGs with XAI models can contribute to more comprehensive decisions through knowledge matching, cross-disciplinary and interactive explanations.  Figure ~\ref{intro_kgfig} describes the overview of our proposed KG and its applications which are demonstrated in this paper. The different elements (SCADA data, XAI model and model predictions) have mutual interaction, and can also consult the KG independently for information querying/retrieval.

We make the KG publicly available, and it can be further developed/expanded in the future as new terms, concepts and maintenance action strategies prevail in the wind industry. We envisage that our work can serve as a building ground for encouraging future research in utilising KGs for O\&M in the wind industry, providing more value and utility to the conventionally opaque AI models to move closer to achieving human-level intelligence.

The paper is organised as follows: Section ~\ref{data_sec} describes the datasets used for constructing the KG and demonstration of its application on an operational turbine. Section ~\ref{kg_databases} introduces KG databases and their basic concepts. The development of the proposed KG is described in Section ~\ref{dev_kg}. Section ~\ref{use_kg} discusses potential use-cases of the KG in decision support. A brief discussion on the experimental results and the wider applicability of the KG to other turbines beyond the LDT is presented in Section ~\ref{discussion}.  Finally, Section ~\ref{conclusion} concludes the paper and provides the path for future work.

\section{Data description}\label{data_sec}
We utilised the publicly available Skillwind Maintenance Manual \footnote{Skillwind Maintenance Manual: \url{https://skillwind.com/wp-content/uploads/2017/08/SKILWIND_Maintenance_1.0.pdf} \label{skillwind_reference}} for developing the maintenance actions segment of our KG. The Skillwind project has seen immense success in applications development of games and learning tools to complement and foster skills for training professionals and engineers in an interactive environment as well as providing solid foundations and knowledge to enhance industrial skills in the wind energy sector \cite{ASTIASOGARCIA2020124549}. The Skillwind Maintenance Manual originally contains 179 pages of domain-specific information, consisting of text, images and tables and was first developed to support training and set skill standards for wind turbine technicians and O\&M personnel. There are extensive and comprehensive details of preventive, corrective and predictive maintenance action strategies for different systems such as blades, yaw system, pitch system etc. There were some specific details in this manual such as glossary of common terms and definitions (like downtime, time to repair etc.) which we have not considered in constructing the KG, as such information is already available widely on the internet and in existing studies.  

Besides this data, we utilised a SCADA dataset from the Levenmouth Demonstration Turbine (LDT) - an operational offshore wind turbine rated at 7 MW, along with historical logs of alarm messages and ground truth of failures in various Functional Groups (sub-components) to support the development of our KG. Note that given the confidential nature of SCADA datasets, we have provided all information which is presently publicly available in our openly released KG, and more details and relevant information is available from the Platform for Operational Data (POD)\footnote{Disseminated by ORE Catapult: \url{https://pod.ore.catapult.org.uk} \label{orec_reference}} and \cite{windenergy_journal}. We utilise 102 SCADA features (such as pitch angle, gearbox oil temperature, wind speed, rotor speed etc.) at generally 10 minutes intervals from the LDT data, with labelled records of 13 categories of anomalies (14 including normal operation) in different turbine sub-components (gearbox, pitch system, yaw system, hydraulic system etc.) and 26 different classes of alarm messages for our study. We integrate the SCADA feature types, fault categories and alarm types with the corresponding maintenance actions in the Skillwind manual. The LDT SCADA dataset is later utilised in the paper, to demonstrate the application of the proposed KG for intelligent and explainable decision support during turbine operation.

\section{Knowledge graph databases}\label{kg_databases}
Knowledge graphs (KGs) are typically semantic networks, which can extract all available information (knowledge) to construct graph structures, wherein, the nodes denote entities in the graph and edges represent the associated relationships between nodes \cite{Yuan_2018}. These networks are commonly referred to as property graphs \cite{hogan2020knowledge}, and can contain property-value pairs in each node, providing conceptual descriptions of domain-specific events and routines. 

Presently, there are several graph database management systems which are commonly used for development of KGs and information retrieval, such as Neo4j, GraphDB, AlegroGraph etc.  \cite{rs70709473}. We chose to utilise Neo4j for developing the KG in this paper, as it is an open-source NoSQL database, has a robust architecture and is highly computationally efficient in terms of its read and write performance. Moreover, Neo4j supports ACID (Atomicity,  Consistency, Isolation and Durability), ensuring reliable processing of transactional operations in the database \cite{Lpez2015LiteratureRA}, which is integral for information management in safety-critical domains like wind energy. Neo4j also consists of a Developer Graph Apps Library, which gives the user the ability to install several specialised applications directly within Neo4j Desktop such as Halin (for cluster-enabled monitoring of live queries and metrics), Graph Data Science Library (for interactive exploration and insights from the KG through graph data analytics), Graphlytic Desktop (for graph analytics and visualisation through HTML5 directly in the web browser) and several others. Such features and provisions make Neo4j an ideal tool for both, professionals and non-technical users in the wind industry.

\begin{figure}[!h]
\centerline{\includegraphics[width=0.7\textwidth]{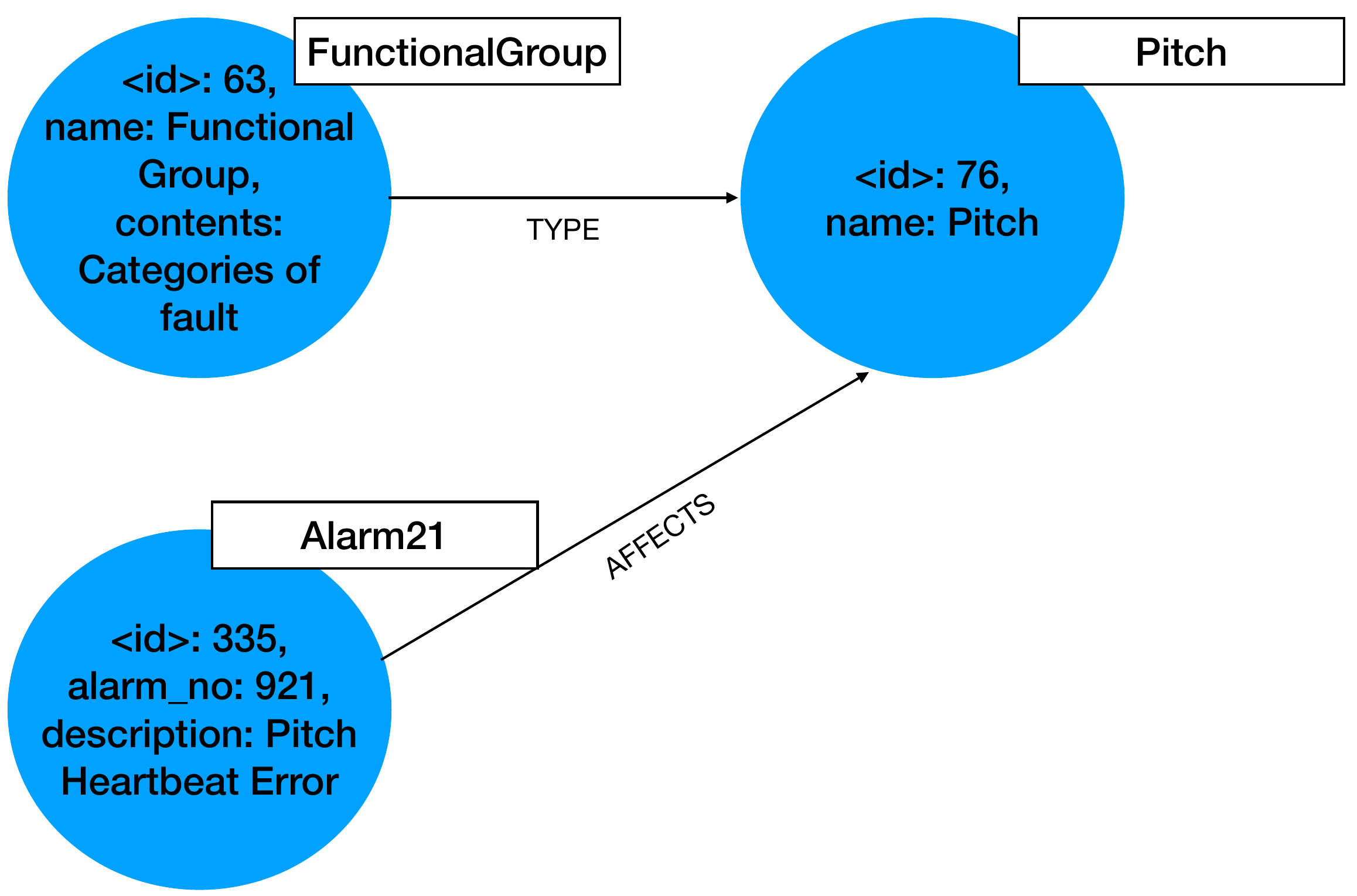}}
\caption{Example description of a KG with 3 distinct nodes, 2 edges and 3 labels defining an alarm in the Pitch system \label{kg_basicsfig}}
\end{figure}

KGs can be directed (wherein, a directional relationship is specified from one node to another) or undirected (with the graph edges not having a specified directional relationship to the nodes). Figure ~\ref{kg_basicsfig} shows a labelled directed KG model, wherein, there are 3 distinct nodes describing the effect of an alarm (\textit{Pitch Heartbeat Error}) on the wind turbine's \textit{Pitch} system. Each node has some properties/attributes such as \textit{name}, \textit{description} etc., which contain more detailed information on the event/phenomenon described by the node. It is possible for different nodes to have the same property labels such as \textit{name}, but their contents will always be distinct and unique. Note that every node has a unique ID assigned automatically by Neo4j to help distinguish it from other nodes. 

\section{Development of the knowledge graph}\label{dev_kg}
\subsection{General structure}
At the most basic level, we consider the wind turbine as a system representing the root node for our KG. The system node is referred to by the name \textit{Study Turbine}, and has properties describing its key specifications, such as rated power of the turbine, its location etc. The key attributes for the LDT specified for this node are shown in Listing ~\ref{sysgraph}. The node can be referred to either using its label, which in this case is \textit{System} or its properties such as \textit{name}, \textit{rated\_power} etc.

  \begin{lstlisting}[float,belowskip=0pt,caption=Description of root node pertaining to the study turbine,label=sysgraph,basicstyle=\small,frame=single,breaklines=True,  xleftmargin=.2\textwidth, xrightmargin=.2\textwidth]
{
  "identity": 15,
  "labels": [
    "System"
  ],
  "properties": {
"name": "Study Turbine",
"location": "Levenmouth,Fife",
"rated_power": "7MW",
"type": "Offshore"
  }
}

\end{lstlisting}

\begin{figure}[!h]
\centerline{\includegraphics[width=0.7\textwidth]{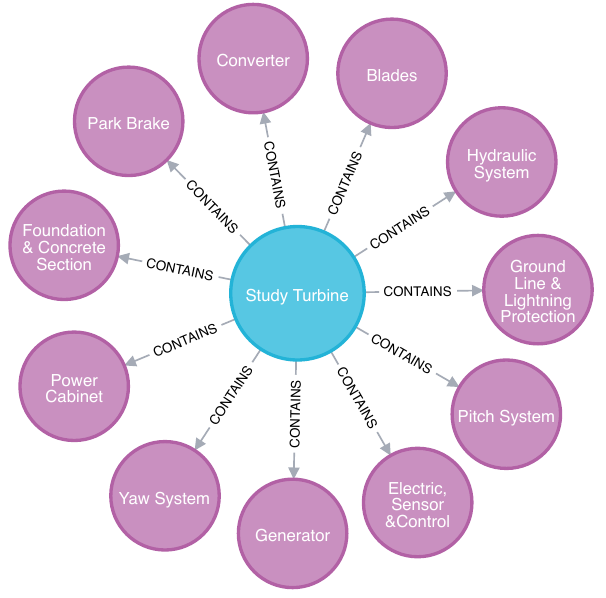}}
\caption{System-level entities for the wind turbine in the KG \label{syslevel_graph}}
\end{figure}  

Based on the Skillwind manual, we created 11 child nodes for the \textit{Study Turbine}, which refer to the different sub-systems wind turbines are generally comprised of (such as pitch system, yaw system etc.). We utilised the \textit{CONTAINS} relationship to interlink these sub-systems with the parent turbine system node. Note that these child nodes are not specific to the LDT, but represent the general constituents of any wind turbine. Figure ~\ref{syslevel_graph} shows the system-level visualisation of the KG. Note that given the large numbers of nodes and their associated relationships in our study, it is infeasible to visualise all nodes together in the size constraints of the figures. We show a subset of relevant nodes in our graph visualisations in this paper. The interactive graph is publicly available on GitHub \footnote{XAI4Wind Supplementary Resources: \url{http://github.com/joyjitchatterjee/XAI4Wind} \label{xai4wind_resources}}, and the user can zoom in/out, view relevant nodes and relationships and perform appropriate query operations easily by using Neo4j. 

For each of these sub-system nodes, we created properties for preventive and corrective maintenance activities which are common and generic to the broad area of these sub-systems. These properties are referred to by \textit{PreventiveActivities} and \textit{CorrectiveActivities} respectively, and encompass a list of multiple actions which are relevant to O\&M of the sub-system. We also created an \textit{InspectionActivities} attribute, describing O\&M strategies for sub-systems like \textit{Blades}, which generally require an external evaluation of the damage prior to maintenance. In addition, wherever available, we included images relevant to the maintenance activities for these sub-systems in the \textit{image\_url} property, wherein, the clickable link directly points the user to the appropriate image stored in our Github repository. Listing ~\ref{transformer_kgexample} shows an example snippet describing the \textit{Transformer} node.

\begin{lstlisting}[float,belowskip=0pt,caption= Example description of the Transformer sub-system, 
  label = transformer_kgexample,
  basicstyle=\small, frame=single,breaklines=True]
{
  "identity": 405,
  "labels": [
    "Transformer"
  ],
  "properties": {
"name": "Transformer",
"image_url": [
      "https://github.com/joyjitchatterjee/XAI4Wind/blob/master/images_maintenance/Transformer_Diagram.png"....
    ],
"CorrectiveActivities": [
      "The transformer could have two different corrective maintenance operations:
1. Replacement in case of failure.
2. Centering coils in case moves are observed.
The procedure of centering coils is made in order to maintain the same distance between
the three phases and to check that LV and HV coils are concentric.",
      "In order to do that procedure it is necessary to ground the transformer and to lock some
disconnections in accordance with the instruction manual consignment. ......"
    ]
  }
}
 \end{lstlisting}

\subsection{Incorporation of the LDT data}
Based on the LDT data, we created a \textit{HAS} relationship between the available \textit{SCADA dataset} and the Functional Group which comprises the data. The Functional Group node has 14 different child nodes (such as Gearbox, Yaw etc.), which were represented using \textit{TYPE} relations. \textit{NoFault} is a special node which we created to denote normal operation of the LDT, wherein, there is no anomaly/operational inconsistency in any of the turbine sub-components. Each type of Functional Group was assigned a unique label and a class number. See Listing ~\ref{pitchfg_des} for an example description of the \textit{Pitch System Interface Alarms} Functional Group. 

  \begin{lstlisting}[float,belowskip=0pt,caption= Example description of the Pitch System Interface Alarms Functional Group, 
  label = pitchfg_des,
  basicstyle=\small, frame=single,breaklines=True]
{
  "identity": 66,
  "labels": [
    "PitchInterfaceAlarm"
  ],
  "properties": {
"name": "Pitch System Interface Alarms",
"fno": 2
  }
}
\end{lstlisting}

\begin{figure}[!h]
\centerline{\includegraphics[width=0.7\textwidth]{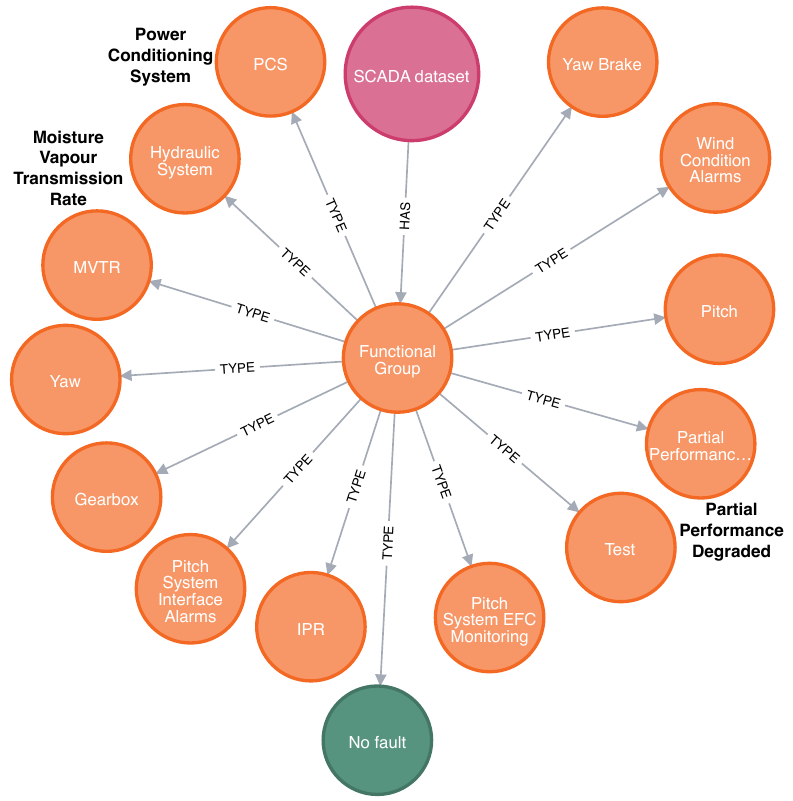}}
\caption{Constituent Functional Groups in the SCADA dataset for the LDT\label{fg_graph}}
\end{figure}  

Figure ~\ref{fg_graph} shows the different constituent Functional Groups in our KG. Note that these nodes are specific to the LDT based on our utilised data, and can have some variations depending on the wind turbine types and their technical specifications. The interested reader can refer to \cite{torque_paper} for more detailed information on these Functional Groups. We mapped these Functional Groups to the generic sub-system nodes of the \textit{Study Turbine} node through \textit{RELATESTO} relationships. As different Functional Groups and sub-systems can themselves be composed of several other sub-components (e.g. drive train in turbines are made up of gearbox and generator), we created \textit{CONSISTSOF} relationships to interlink them.

Based on the 102 SCADA features we had in the dataset, we also created 102 distinct nodes representing these features. Each node is of the label type \textit{FEATURE}, and has attributes describing the feature like name, measurement units and a brief description of the feature. Listing ~\ref{gbox_feature_des} shows an example describing a SCADA feature relating to the turbine gearbox. 

  \begin{lstlisting}[float,belowskip=0pt, caption= Example description of a SCADA feature pertaining to the turbine gearbox, 
  label = gbox_feature_des,
  basicstyle=\small, frame=single,breaklines=True]
{
  "identity": 57,
  "labels": [
    "Feature"
  ],
  "properties": {
"name": "GBoxOpShaftBearingTemp2_Stdev",
"description": "Gearbox Bearing 2 
Temperature Standard Deviation",
"unit": "Deg celsius",
"feature_no": 46
  }
}
 \end{lstlisting}

We created 26 different nodes representing different alarm types in the LDT, wherein, each alarm is described with a brief description and has a unique \textit{alarm\_no} attribute in the range \textit{901-926}. See Listing ~\ref{blade_alarmdes} for an example snippet describing an alarm in the turbine's \textit{Blades}. Due to the confidential nature of alarm logs in the wind industry, we have provided some examples (alarms \textit{Alarm1} to \textit{Alarm5}) of these in our publicly released KG. More information on these can be found at \cite{9206839}.

  \begin{lstlisting}[float,belowskip=0pt,caption= Example description of an alarm for turbine blades, 
  label = blade_alarmdes,
  basicstyle=\small, frame=single, breaklines=True]
{
  "identity": 318,
  "labels": [
    "Alarm4"
  ],
  "properties": {
"description": "Blade 1 too slow to respond",
"alarm_no": "904"
  }
}
 \end{lstlisting}

\subsection{Mapping fault events in the Skillwind manual to the LDT Functional Groups}
\begin{figure}[!h]
\centerline{\includegraphics[width=0.8\textwidth]{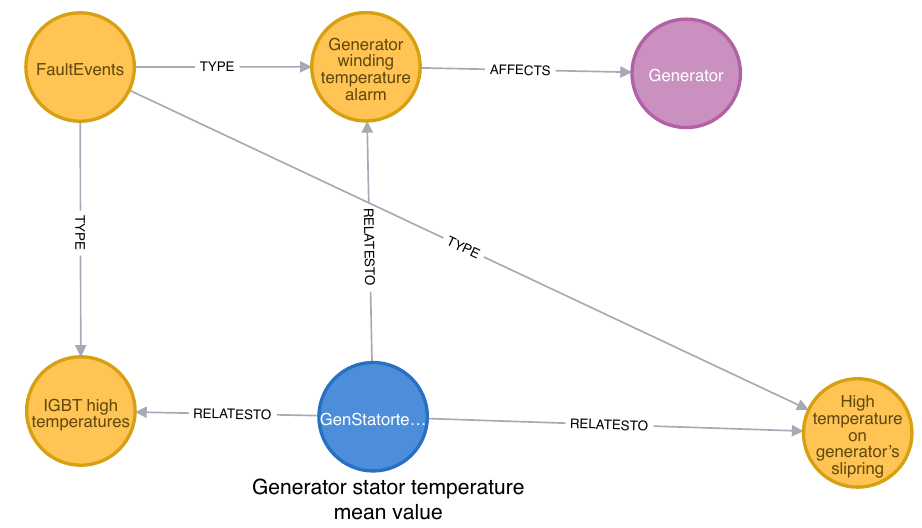}}
\caption{Fault event sub-graph for an anomaly in the Generator \label{faultevent_graph}}
\end{figure}

Based on the Skillwind manual, we had a collection of 57 different types of fault events which are common to any wind turbine, and we created 57 separate nodes denoting the same, which are basically child nodes of \textit{FaultEvents} with \textit{TYPE} relations. Each fault event node \textit{FaultEvent1} to \textit{FaultEvent57} has a \textit{details} attribute, briefly describing the nature of the fault. We mapped the 26 different alarms to the corresponding fault events with \textit{RELATESTO} relations. As specific SCADA features can give rise to anomalies in the turbine sub-components \cite{windenergy_journal}, we also created \textit{RELATESTO} relationships between the relevant features which contribute to the fault events (and thereby lead to the alarms). Figure ~\ref{faultevent_graph} enunciates an example sub-graph for fault events affecting the \textit{Generator}, wherein, the \textit{Generator stator temperature mean value} SCADA feature (attributing a high temperature in the generator stator winding) is related to the fault events, leading to the \textit{Generator winding temperature alarm} in the LDT. 

\subsection{Operations \& maintenance actions for different sub-systems}
As decisions pertaining to O\&M of wind turbines can vary widely owing to the presence of multiple complex scenarios and a large number of variables \cite{en12020225}, preventive, predictive and corrective actions can be unique to each sub-system in the turbine, or it may also be possible for similar actions to fix/avert different types of failures \cite{9206839}. We created a specific node referenced with the label \textit{MaintenanceAction}, which has 3 child nodes connected with \textit{ACTION} relationships: \textit{Preventive}, \textit{Predictive} and \textit{Corrective} representing corresponding maintenance actions respectively. We also added several properties into the \textit{Preventive} node, which contain maintenance actions generally common to any component in the turbine such as \textit{Cleaning} activities, \textit{Retightening} actions etc. Listing ~\ref{preventivenode} shows some examples of these attributes. 

  \begin{lstlisting}[float,belowskip=0pt,caption= Example description of properties in the Preventive node,label = preventivenode, basicstyle=\small, frame=single, breaklines=True ]
{
  "identity": 79,
  "labels": [
    "Preventive"
  ],
  "properties": {
"Cleaning": [
      "Cleaning everything (grease, oil, dust, rags
      ,other maintenance, carbon rings, collectors, etc.)",
      "Treating waste in accordance with procedures .....
    ],
......
"Retightening": [
      "All bolted joints need to be 
      proven to not lose their tightening."
    ]
    ]
  }
}
 \end{lstlisting}

\begin{figure}[!h]
\centerline{\includegraphics[width=0.8\textwidth]{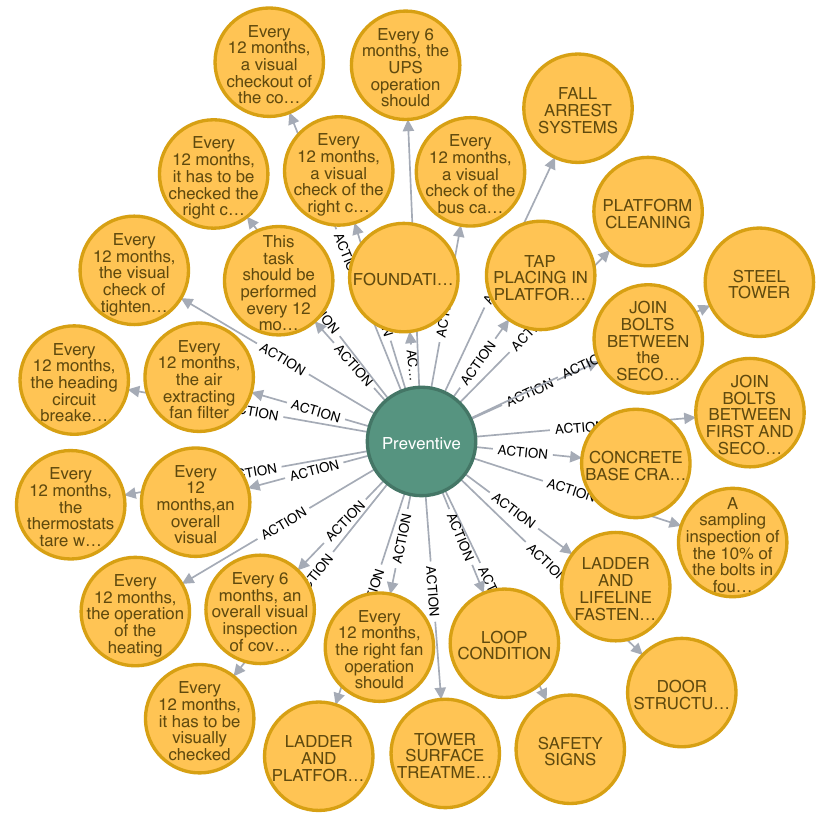}}
\caption{Example visualisation of few nodes for preventive maintenance actions \label{subset_preventive}}
\end{figure}

\paragraph{\textbf{Preventive maintenance activities}}
For preventive actions which are specific to any of the turbine sub-systems (or Functional Groups), we created 233 distinct nodes referenced with labels from \textit{PrevAct1} to \textit{PrevAct233}. We also created several properties for each node, such as \textit{name} (describing the activity briefly), \textit{gen\_periodicity} (general periodicity of the action), \textit{activities} (thorough and comprehensive description of the action(s)), \textit{initial\_periodicity} (initial periodicity of the action) and \textit{act} (numeric reference to the action). To explicitly specify the preventive actions which are required for individual sub-components, we created \textit{FOR} relationships to map such actions to the corresponding sub-systems. A subset of all preventive maintenance actions created in the KG is visualised in Figure ~\ref{subset_preventive}. 

\begin{figure}[!h]
\centerline{\includegraphics[width=0.8\textwidth]{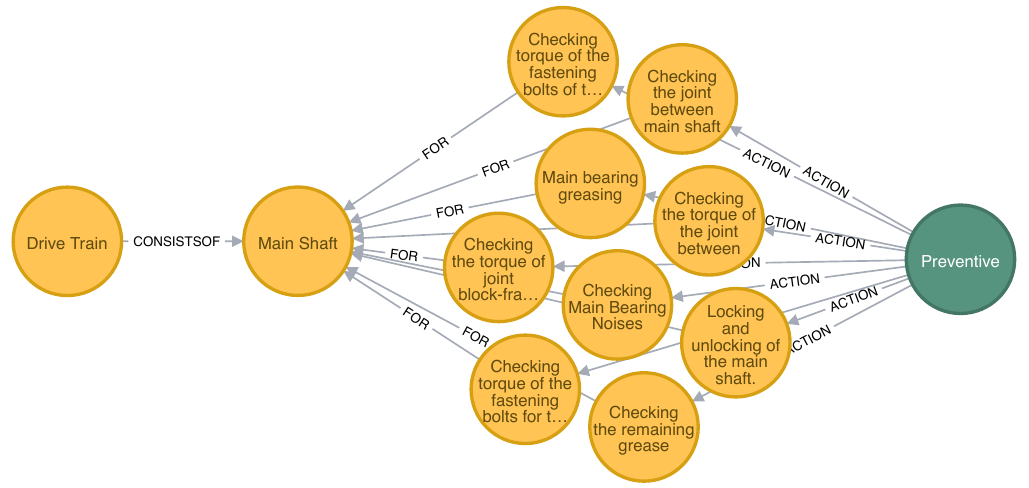}}
\caption{Example of preventive maintenance actions required for the main shaft\label{prev_eg}}
\end{figure}

Figure ~\ref{prev_eg} shows an example of multiple preventive actions for the \textit{Main Shaft} in the turbine (which is a part of the \textit{Drive Train}). Further, each of these nodes representing preventive actions contains further details within the node's attributes, see Listing ~\ref{mainbear_eg} for an example. 

  \begin{lstlisting}[float,caption= Example of detailed description for a preventive action for main shaft, 
  label = mainbear_eg,
  basicstyle=\small, frame=single,breaklines=True]
{
  "identity": 181,
  "labels": [
    "PrevAct100"
  ],
  "properties": {
"details": "Checking Main Bearing Noises",
"act": 100,
"activities": [
      "It is mandatory to listen to any noise or vibration from the bearing mounting when rotor is turning slowly.",
      "Another verification is to make the rotor turn slowly, then stop the turbine by pushing an emergency button and looking at the clearance between the main shaft and the bearing shield."
    ]
  }
}
 \end{lstlisting}
 
\begin{figure}[!h]
\centerline{\includegraphics[width=0.8\textwidth]{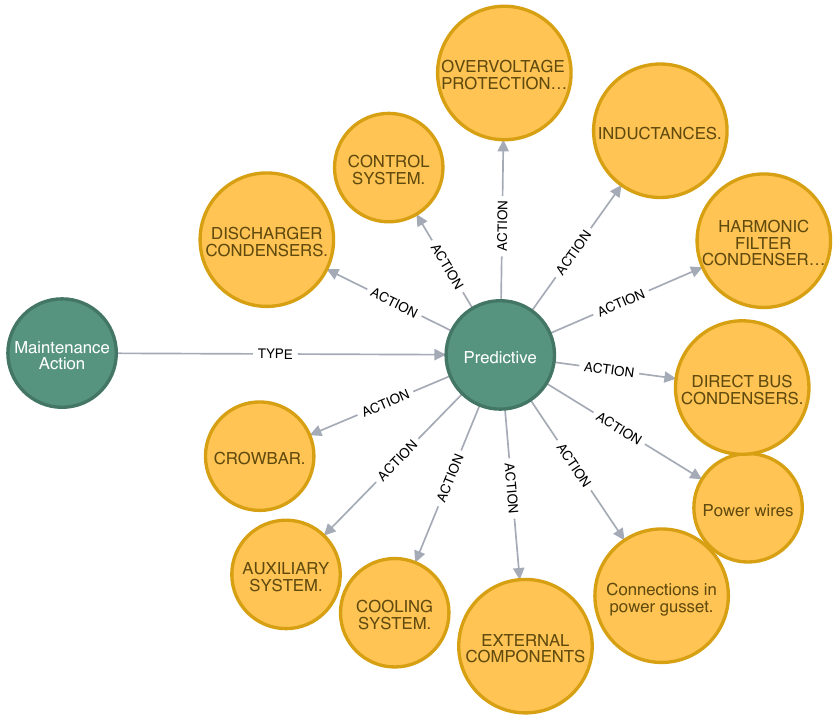}}
\caption{Different types of predictive actions in the knowledge graph \label{pred_types}}
\end{figure}

\begin{figure}[!h]
\centerline{\includegraphics[width=0.8\textwidth]{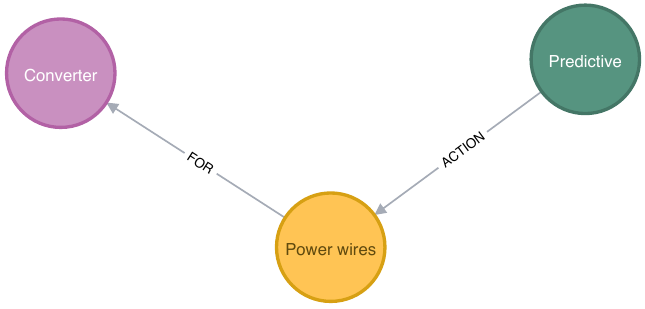}}
\caption{Visualisation of the predictive maintenance action for the converter \label{pred_converter_eg}}
\end{figure}

\paragraph{\textbf{Predictive maintenance activities}}
We created 11 distinct nodes representing 11 distinct maintenance actions based on the Skillwind manual, referenced by the labels \textit{PredAct1} to \textit{PredAct11}. Similar to the nodes for preventive maintenance actions, the predictive action nodes were assigned attributes like \textit{details} (brief description of the maintenance activity), \textit{activities} (list of multiple predictive actions), \textit{image\_url} (links to image(s) representing the action required) etc. We also created \textit{FOR} relationships between these action nodes with the sub-components/sub-systems which are relevant to their condition monitoring. Figure ~\ref{pred_types} depicts the various types of predictive maintenance actions which were created for different assets in the turbine.  An example description of a predictive action for the \textit{Converter} is shown in Listing ~\ref{predtype_kg}, with a visualisation of the relationship in Figure ~\ref{pred_converter_eg}.

  \begin{lstlisting}[float,caption= Example description of a predictive action for power wires of the converter, 
  label = predtype_kg,
  basicstyle=\small, frame=single,breaklines=True]
{
  "identity": 528,
  "labels": [
    "PredAct1"
  ],
  "properties": {
"details": "Power wires",
"activities": [
      "The power connections are formed by wires through which high intensities run, converter input and output and its rotor exits, connected with ring connectors through whose hole passes the screw that is to join them to the terminal passes.",
     .....
    ],
"image_url": [
      "https://github.com/joyjitchatterjee/XAI4Wind/blob/master/images_maintenance/Wires_MountedOverReactance.png"....
    ]
  }
}
 \end{lstlisting}
 
\begin{figure}[!h]
\centerline{\includegraphics[width=0.8\textwidth]{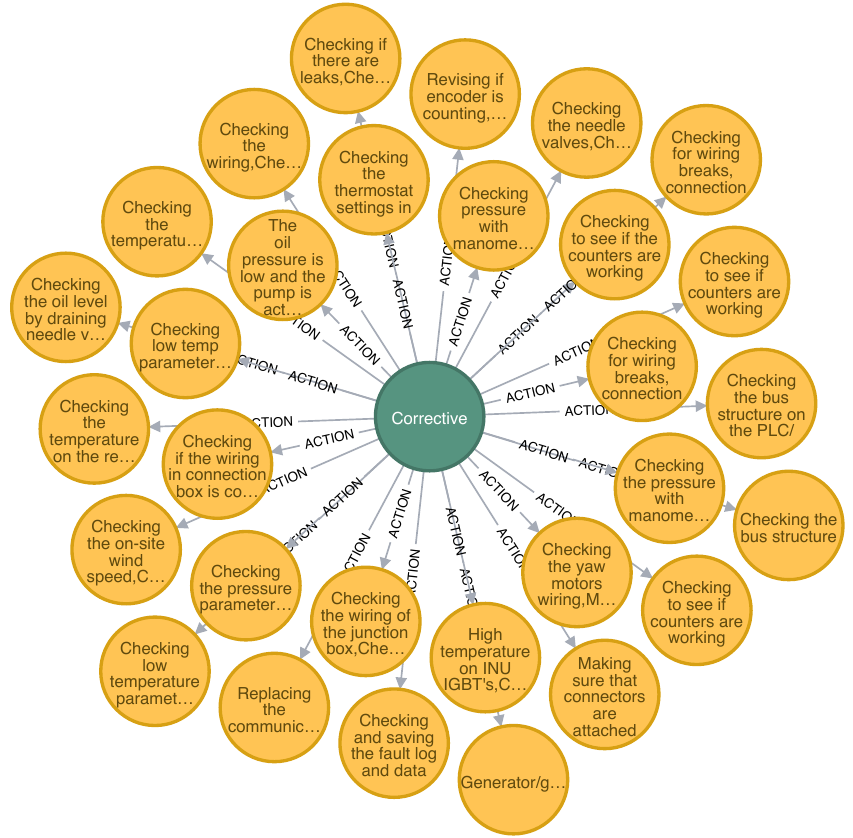}}
\caption{Example of few types of corrective actions \label{correctivetype_eg}}
\end{figure}

\paragraph{\textbf{Corrective maintenance activities}}
For incorporating correcting maintenance actions into the KG, which are specific to either of the 57 fault events (which in-turn lead to an associated alarm event), we utilised the Skillwind manual to create 57 distinct corrective action nodes referenced by the labels \textit{CorrAct1} to \textit{CorrAct57}. For each of these nodes, we also created the attributes \textit{activities} (representing a list of multiple corrective action(s)) and \textit{image\_url} (link to the appropriate image representative of the action). Figure ~\ref{correctivetype_eg} shows a subset of all corrective actions as an example.

\begin{figure}[!h]
\centerline{\includegraphics[width=0.8\textwidth]{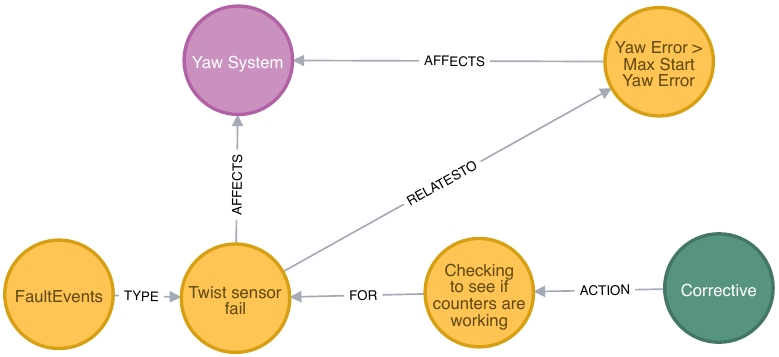}}
\caption{Visualisation of corrective action required for twist sensor failure in the yaw system \label{correctivets_eg}}
\end{figure}

Similar to preventive and predictive actions, we created \textit{FOR} relationships between the corrective action nodes and the fault events which these actions help to fix. An example of corrective actions required for a \textit{Twist sensor fail} fault event in the \textit{Yaw System} is depicted in Figure ~\ref{correctivets_eg}. A detailed example description of the corrective action node is provided in Listing ~\ref{correctivets_kg}.

\begin{lstlisting}[float,caption= Example describing corrective maintenance action for an alarm in the yaw system, 
  label = correctivets_kg,
  basicstyle=\small, frame=single,breaklines=True]
  
{
  "identity": 410,
  "labels": [
    "CorrAct1"
  ],
  "properties": {
"activities": [
      "Checking to see if counters are working    correctly in manual yaw. If not, the encoder
      has to be replaced",
      "Untwisting the turbine, resetting turtle, and resetting the counters on the touchscreen",
      "Checking wiring breaks, connection points, cuts",
      "Checking connection on the I/O card"
    ]
  }
}
 \end{lstlisting}

It is integral to note from this example that for all types of maintenance actions (preventive, predictive and corrective) incorporated in our KG, there exist several other associations and relationships with other nodes, such as the alarm type (\textit{Yaw Error > Max Start Yaw Error}), Functional Group (\textit{Yaw System}) etc. which have been discussed earlier. We provide the complete details of all nodes in the KG, along with a brief description of their utility and the available properties in  ~\ref{kg_nodes_details}. The presence of multiple relationships and nodes \footnote{Some nodes may not have specific associated relationships (e.g. Communications \& Network Sub-System does not share relationships with SCADA features based on our datasets utilised), but new relationships can be incorporated in the future.} gives rise to ambient intelligence in the proposed KG, and can lead to several real-world applications in the wind industry, which we discuss next.

\section{Use-cases of the proposed knowledge graph}\label{use_kg}
Below, we discuss potential use-cases and real-world applications of the proposed KG in the wind energy domain. Note that there are a plethora of applications which can be leveraged from the KG through the Neo4j apps outlined in Section ~\ref{kg_databases}, but in our discussion below, we focus on 3 specific areas which we believe are most relevant to the wind industry for explainable decision support in O\&M.
\subsection{Querying the graph through Cypher}
The most basic application which can directly be leveraged from the proposed KG is the utilisation of the Cypher programming language implemented in Neo4j to query the KG. At present, the Cypher language for property graph querying is extensively used for development and management of commercial databases and products in industry, and by researchers  \cite{cypher_paper}. The proposed KG can easily be queried using the common keywords in Neo4j such as \textit{MATCH}, \textit{WHERE}, \textit{RETURN} etc. A simple one-liner in Neo4j (either hosted on a web server such as the Neo4j Sandbox or directly in the desktop application) can be used to retrieve relevant nodes and relationships, including their visualisations and detailed descriptions. 
The interested reader can refer to relevant papers such as \cite{cypher_paper,guia_soares_bernardino_2017} and the official Neo4j documentation \footnote{Neo4j Documentation: \url{https://neo4j.com/docs/} \label{neo_docs}} for more details on Cypher and its syntax.

\begin{figure}[!h]
\centerline{\includegraphics[width=0.8\textwidth]{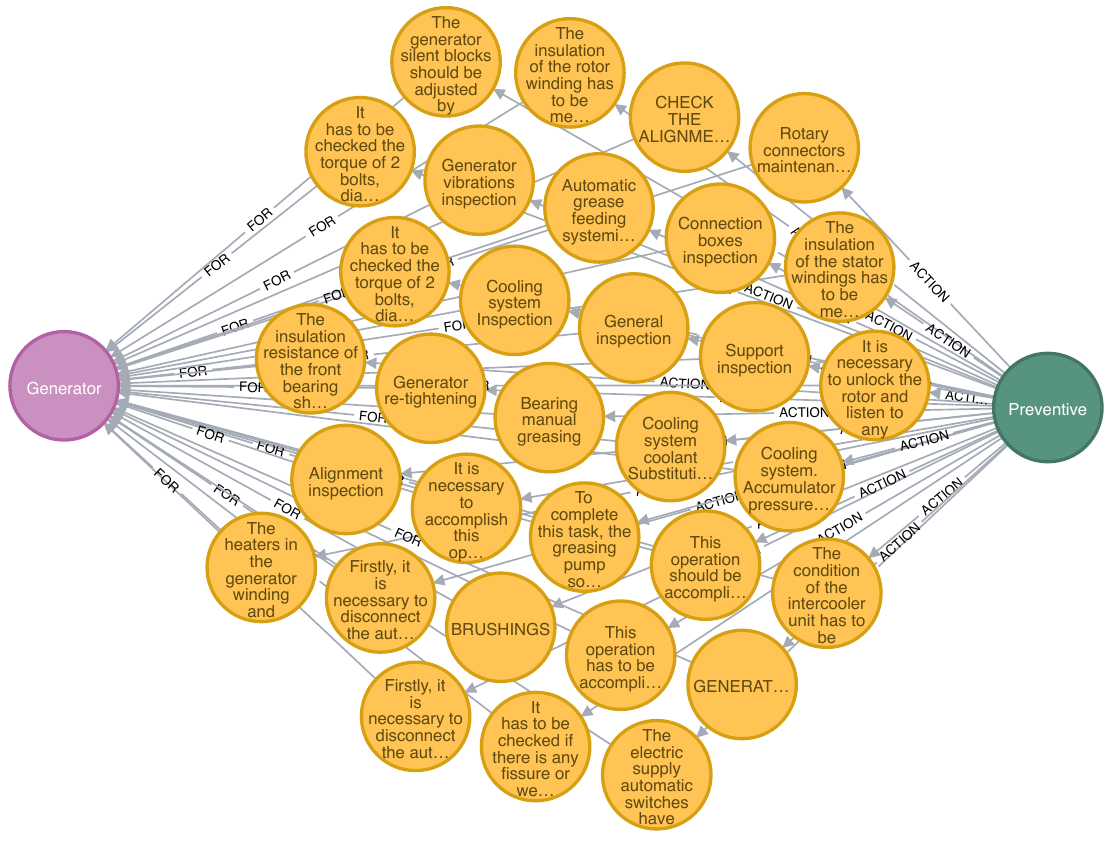}}
\caption{Visualisation of the end result of query in preventive actions for the generator \label{match_geneg}}
\end{figure}

As an example of querying a Neo4j KG, to retrieve preventive maintenance actions for the \textit{Generator}, the one-liner code below can be utilised:

\begin{verbatim}
MATCH(n:Preventive)-[:ACTION]->(p)-[:FOR]->(q:Generator) RETURN n,p,q
\end{verbatim}

The end result of this query is shown in Listing ~\ref{geneg_kg} providing an example description of one of the required preventive actions. The resultant graph following this query is shown in Figure ~\ref{match_geneg}, and can provide useful and easy-to-comprehend information to the end user in facilitating decision support.

  \begin{lstlisting}[float,caption= End result towards an example preventive maintenance action query for Generator, 
  label = geneg_kg,
  basicstyle=\small, frame=single,breaklines=True]
{
  "identity": 276,
  "labels": [
    "PrevAct195"
  ],
  "properties": {
"details": "WINDING HEATERS",
"act": 195,
"activities": [
      "Firstly, it is necessary to disconnect the automatic switch. Then, the resistance connected would be released and it would be checked that the resistor is not open. If it is open, it must be changed."
    ]
  }
}
 \end{lstlisting}
 
Despite its advantages, the use of Cypher has the key drawback that it requires basic technical know-how on part of the wind turbine engineers \& technicians in Cypher programming within Neo4j and knowledge graph databases. We believe that there are better alternatives to facilitate automated decision making instead of solely performing syntactic queries, as discussed next.
\subsection{Interactive graph exploration in Neo4j Bloom}
Neo4j contains a specialised graph exploration and visualisation tool called Bloom \footnote{Neo4j Bloom: \url{https://neo4j.com/bloom/} \label{bloom_ref}} , which facilitates querying the KG with support for custom search phrases in natural language, interactive visualisation/exploration and discovery of novel patterns and insights based on context of the retrieved information. Bloom facilitates the information available in the KG to be analysed and investigated visually from varying perspectives, and has a specialised search bar which provides type-ahead suggestions when querying the KG through Cypher commands. 

We believe that the ability of Bloom to provide near natural language search can be significantly beneficial in the wind energy sector. Custom search phrases with human-readable text can be created during the development phase to provide engineers \& technicians with a platform for retrieval of appropriate information without the need for any coding. Note that Bloom supports both static and dynamic search phrases, for which Cypher queries can be pre-defined and mapped to natural language text. Once a query is created, it is automatically stored in the Bloom perspective, and future queries can simply be performed via natural language search phrases. Bloom essentially breaks down the search phrases into multiple word tokens, and utilises the data present in the KG together with its interpretation of the graph schema for information retrieval. Below, we discuss the role of both types of queries for decision support through Neo4j Bloom's user-friendly interactive interface.

\begin{figure}[!h]
\centerline{\includegraphics[width=\textwidth]{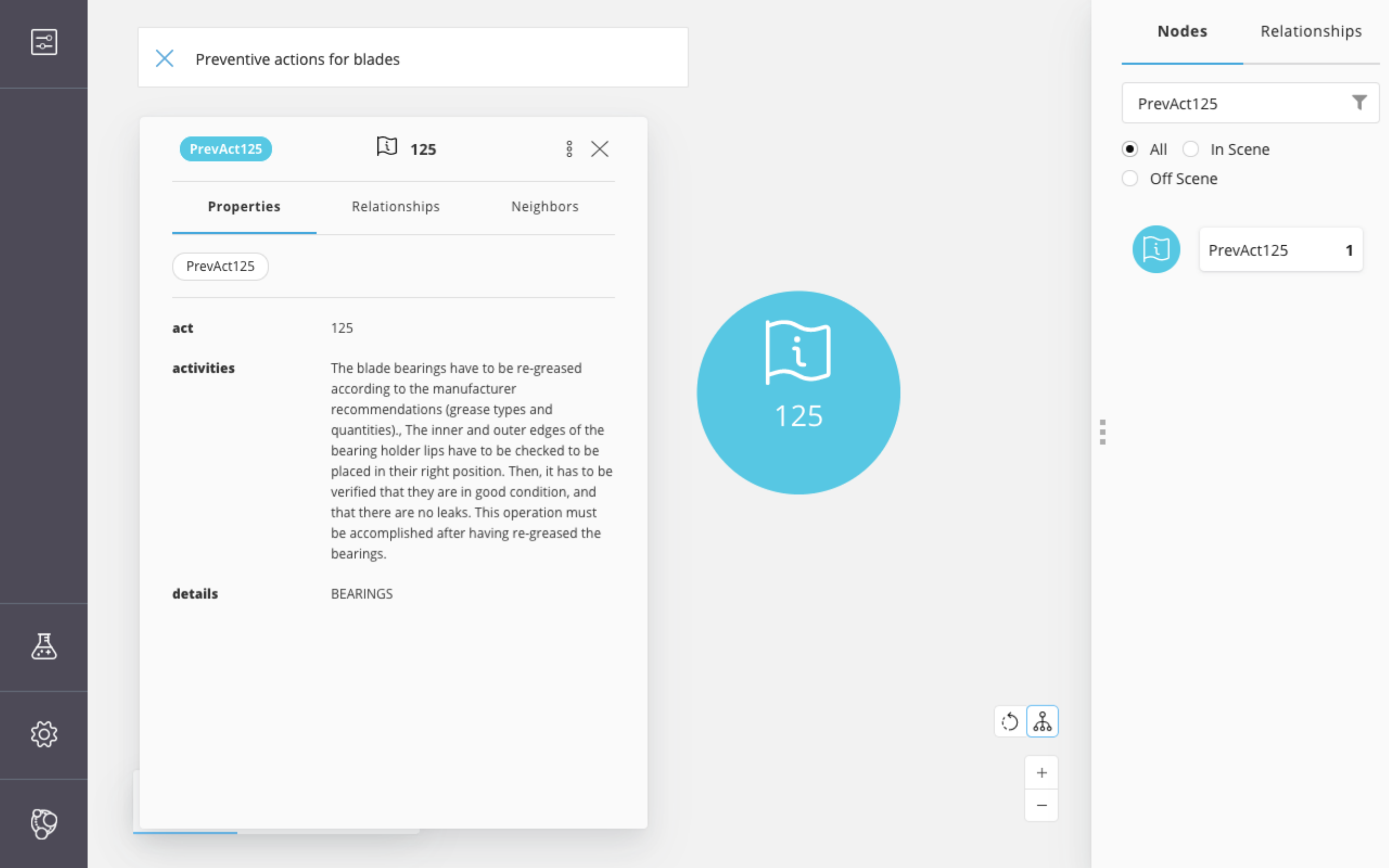}}
\caption{Example screenshot of the Neo4j Bloom interface providing preventive maintenance actions for turbine blades, given corresponding natural language search phrase \label{screenshot_staticblade}}
\end{figure}

\paragraph{\textbf{Static queries in Bloom:}} In Bloom, static queries are the search phrases mapped to Cypher queries which never change and are hard-coded to only retrieve the exact information and records which they have been specifically linked to. An example for creation of a static query for retrieving preventive actions pertaining to the turbine's blades is shown in Listing ~\ref{static_blades}. Note that informative descriptions for the search phrases can also be provided during the development phase, which can help engineers better understand the utility of these search phrases. In this example, a simple natural language search phrase - \textit{Preventive actions for blades} retains multiple nodes pertaining to preventive actions for the blades. Figure ~\ref{screenshot_staticblade} shows an example screenshot of the user-friendly interface in Bloom, which provides the ability to easily type the search phrase and retrieve appropriate O\&M strategies in such scenarios. 

\begin{lstlisting}[float,belowskip=0pt,caption=Example description of creating a static search phrase query in Bloom for preventive maintenance actions for blades,label=static_blades,basicstyle=\small,frame=single,breaklines=True]
Search phrase: Preventive actions for blades
Description: Preventive maintenance actions for the wind turbine's blades
Cypher query:  
MATCH(n:Preventive)-[:ACTION]-(p)-[]-(q:Blades) RETURN *
\end{lstlisting}

\paragraph{\textbf{Dynamic queries in Bloom:}}
Unlike static search phrases, Bloom also supports dynamic queries which can utilise parameters in the query strings to determine the appropriate information which should be retrieved. These search phrases can support multiple parameter data types, such as integers, strings, Boolean values, floats etc., making them powerful tools as a single search phrase can retrieve multiple relevant nodes and relationships. Listing ~\ref{dynamic_phraselist} shows an example for creating dynamic search phrases in Bloom pertaining to retrieving corrective maintenance functions for abnormality in SCADA features. Here, \$description is a dynamic parameter used to match the SCADA feature (e.g. Absolute Wind Direction Mean Value) to the corresponding maintenance actions present in the KG. It is interesting to mention that given that we have 102 SCADA features in this study, dynamic queries help create a single query for corrective actions pertaining to all SCADA features, instead of creating 102 unique/distinct queries which can be a time-consuming and complex task. This can particularly be beneficial in cases wherein there are hundreds or thousands of SCADA features, particularly for large wind farms.

\begin{lstlisting}[float,belowskip=0pt,caption=Example description of creating a dynamic search phrase query in Bloom for corrective actions to fix/avert anomalies due to inconsistency in SCADA features,label=dynamic_phraselist,basicstyle=\small,frame=single,breaklines=True]
Search phrase: Corrective actions for abnormal $description
Description: Corrective maintenance actions for [Choose feature description] abnormal important feature 
Cypher query:  MATCH(n:Corrective)-[:ACTION]->(p)-[:FOR]->(q)-[:RELATESTO]-(r:Feature) WHERE r.description = $description
RETURN  *
\end{lstlisting}

The \$description parameter is a \textit{string} type variable, which basically maps the \textit{description} attribute of the SCADA feature in the KG to the corresponding natural language search phrase's user-provided attribute by executing a Cypher query in the back-end as shown below.
\begin{verbatim}
MATCH(n:Feature) RETURN n.description
\end{verbatim}

\begin{figure}[!h]
\centerline{\includegraphics[width=\textwidth]{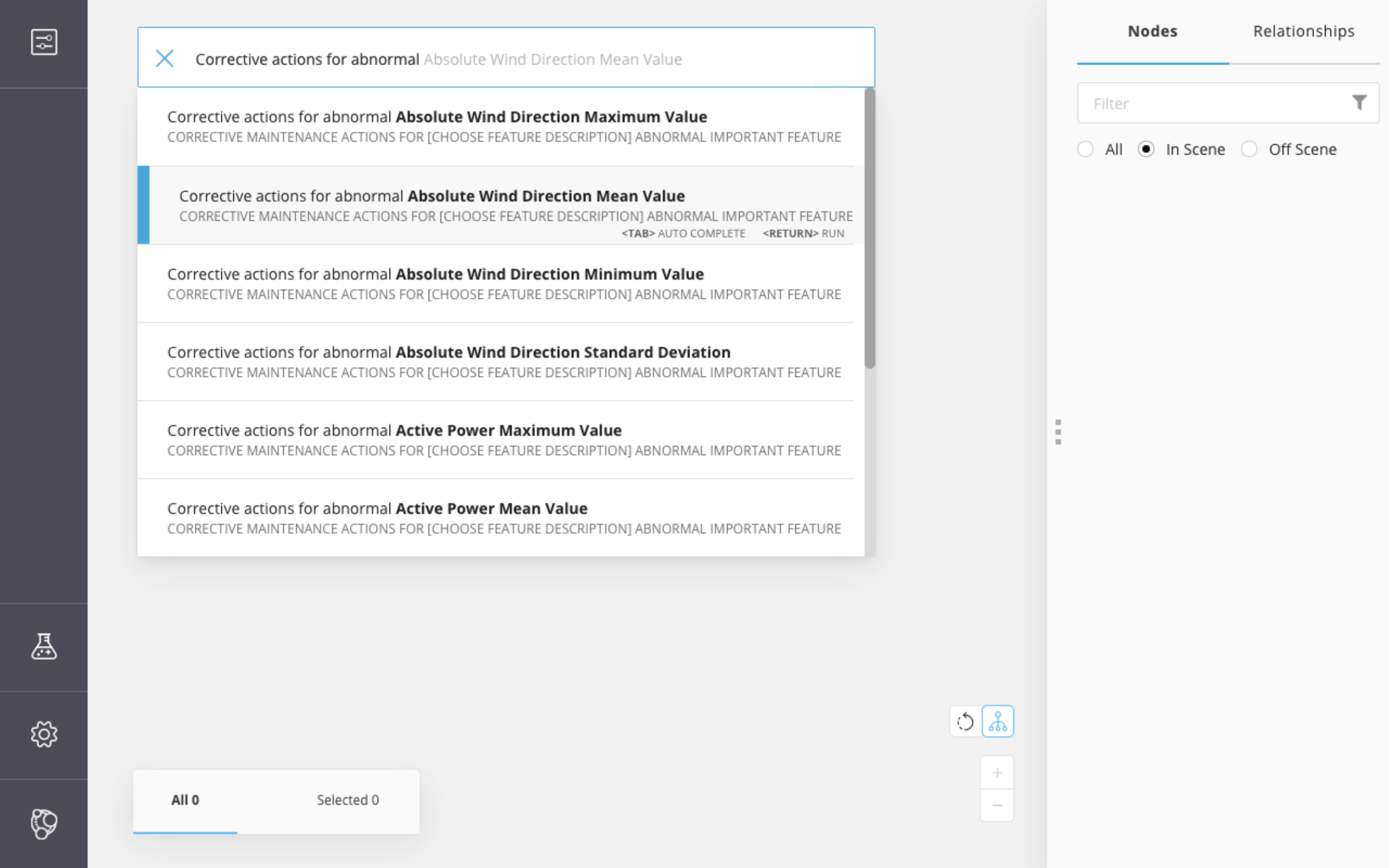}}
\caption{Example screenshot of the Neo4j Bloom interface providing dynamic suggestions for SCADA feature descriptions, given natural language search phrase   \label{suggest_dynamic}}
\end{figure}

\begin{figure}[!h]
\centerline{\includegraphics[width=\textwidth]{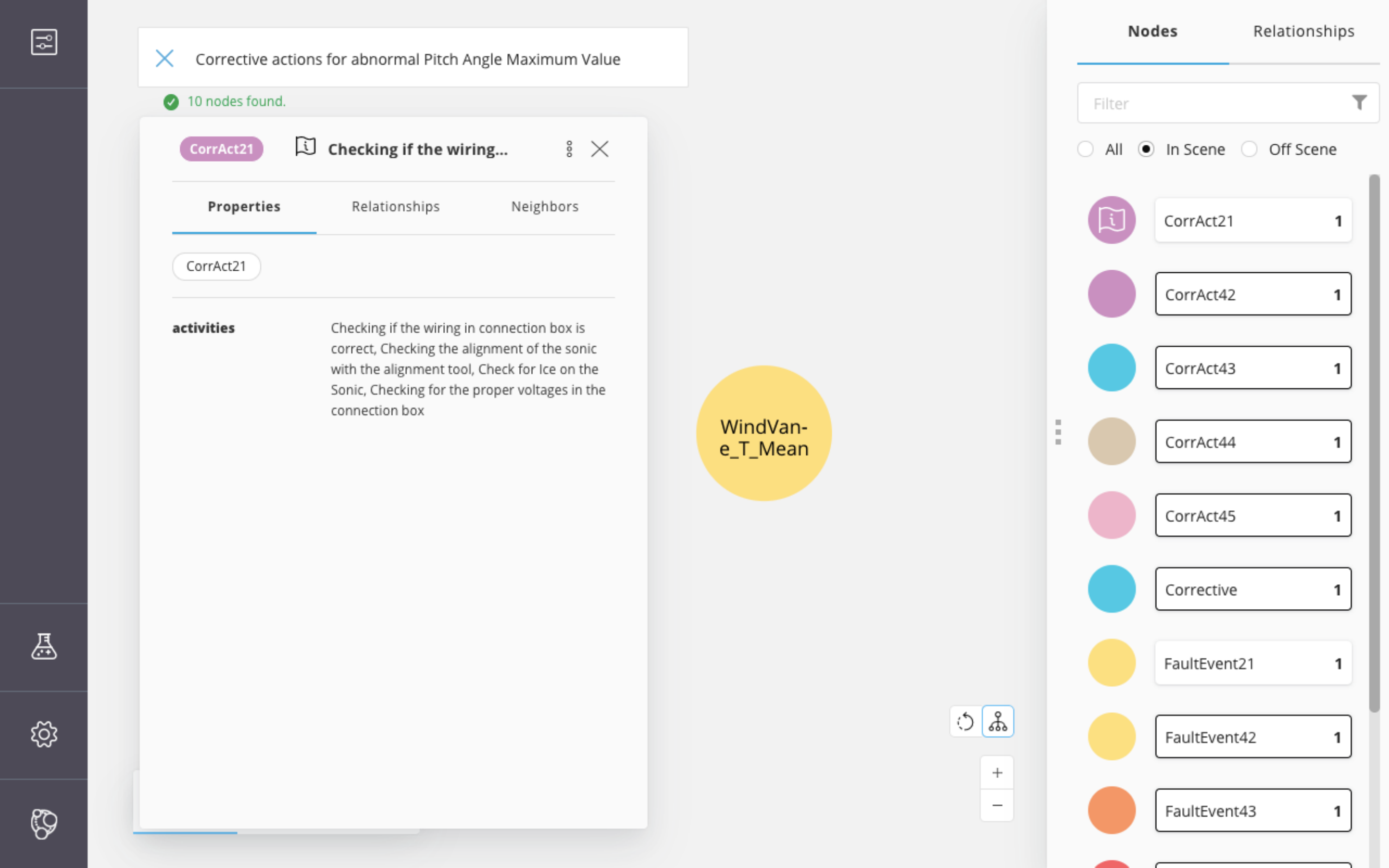}}
\caption{Example screenshot outlining corrective actions for anomalous behaviour of the Pitch Angle Maximum Value SCADA feature\label{screenshot_scadafeature}}
\end{figure}

In the user-friendly Bloom interface, when the search phrase for \textit{Corrective actions for abnormal \$description} is entered, Neo4j Bloom also provides suggestions (in an alphabetical order) for relevant feature descriptions which the user can directly choose from in the interface. Figure ~\ref{suggest_dynamic} shows an example of suggestions provided by Bloom for SCADA feature descriptions based on the KG schema and data. This provision makes Bloom extremely convenient as engineers \& technicians do not need to remember/memorise the exact feature descriptions to retrieve maintenance actions from the KG. An example of corrective maintenance actions required to fix/avert inconsistency in the \textit{Pitch Angle Maximum Value} SCADA feature (which can further contribute to a fault in the pitch system), as obtained through Bloom's interface is shown in Figure ~\ref{screenshot_scadafeature}. Note that there are multiple corrective actions possible in this case (as evident with 10 nodes found in the interface), and we show an example of one such action.

\subsection{Application of graph data science algorithms for insightful reasoning and analytics}
While the KG in itself serves as a powerful resource for information querying and retrieval, in real-world complex industrial use-cases, it is integral to understand the contextual impact of the KG through discovery of key insights, such as the root causes of events and discovery of communities revolving around a common theme \cite{gds_singhdhan}. Graph data analytics can help accomplish this objective, and provide a pathway to the stakeholders for development of new tools and algorithms for optimisation of existing activities. 

Neo4j contains the Graph Data Science Library, which is a native application for graph data analytics. Through a simple and interactive user interface referred to as NEuler, a no-code platform is available, helping users apply graph data science algorithms on datasets. There are several available algorithms built into the platform focused on centrality, community detection, path finding and similarity, which can help in discovering significant novel insights. As an example, we discuss a centrality algorithm applied to the proposed KG which can be of interest to the wind industry:-

\paragraph{\textbf{PageRank for determining influential nodes}} Ranking on KGs is vital for the determination of node importances, with several applications in real-world domains, such as information retrieval, neuroscience, social networks etc \cite{8622563}. PageRank \cite{ilprints422} is a very popular iterative centrality algorithm, which utilises the topological structure of KG to quantitatively measure the transitive influence and interconnection between nodes. It can be utilised to determine influential elements in the KG (like subgraphs, edges and nodes). In the context of the wind industry, identifying influential elements from the proposed KG can help understand the most important metrics for decision support, which can help engineers prioritise events and activities for such elements. 
    
\begin{figure}[!h]
\centerline{\includegraphics[width=\textwidth]{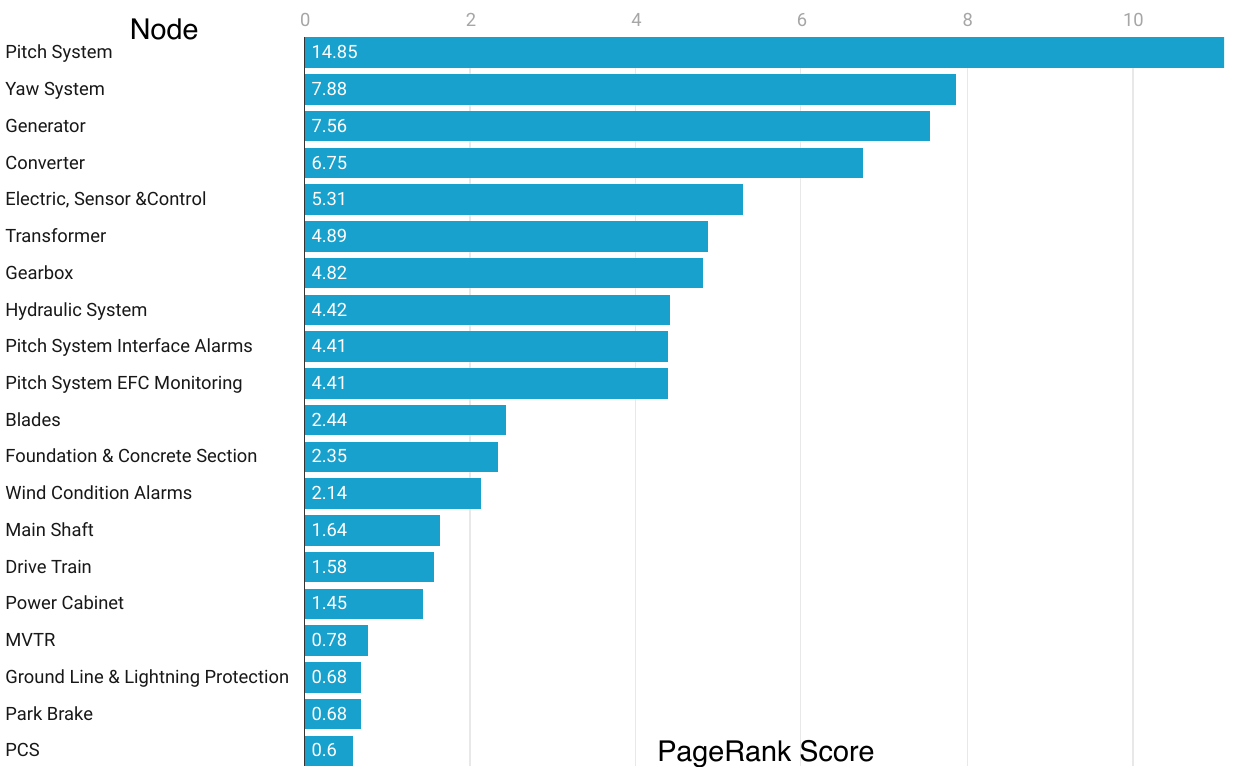}}
\caption{Influence analysis of turbine sub-systems based on PageRank scores\label{pr_subsystems}}
\end{figure}

    We utilise the PageRank algorithm available in the Neo4j Graph Data Science Library for determining influential elements in our proposed KG. For the algorithm, we used 20 iterations, a damping factor of 0.85 and a natural relationship orientation for the projected graph. The algorithm assigns a score to each element in the KG, which provides an approximation of the importance of the element with a higher score signifying a more influential element. Figure ~\ref{pr_subsystems} shows the PageRank scores for various sub-systems and Functional Groups in our KG. As can clearly be seen, the nodes for the \textit{Pitch System}, \textit{Yaw System}, \textit{Generator}, \textit{Electric, Sensor \& Control}, \textit{Transformer} and \textit{Gearbox} are amongst the elements with the highest PageRank scores, signifying their importance for O\&M. This is highly relevant based on existing literature, as these sub-systems generally have the highest frequency of failures determined during structural reliability analysis of wind turbines \cite{en10122099}. This insight from the KG can help engineers \& technicians to especially focus on the important nodes during O\&M, e.g. stocking higher quantity of spare parts for these sub-systems during inventory planning.
    
\begin{figure}[!h]
\centerline{\includegraphics[width=1.3\textwidth,height=0.5\textheight]{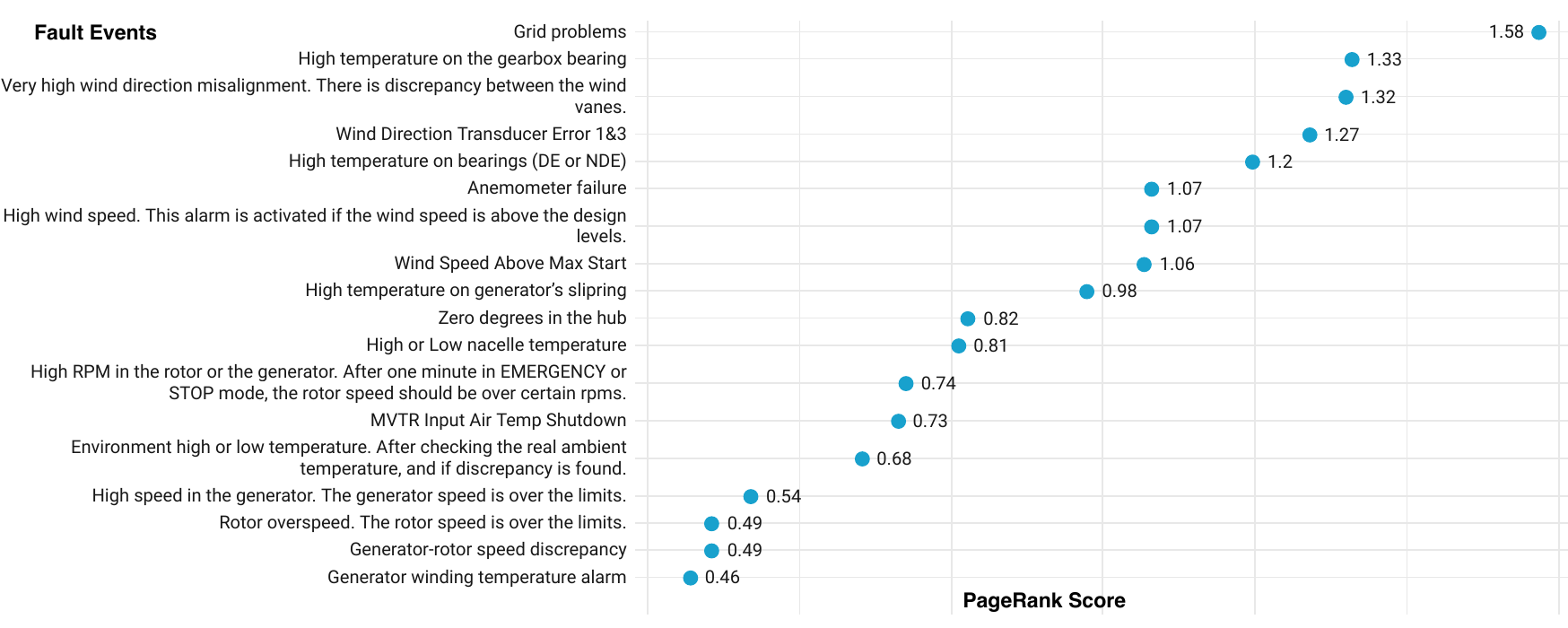}}
\caption{Influence analysis of fault events based on PageRank scores\label{pr_alarmevents}}
\end{figure}

Similar to PageRank scores obtained for the importance of sub-systems, novel insights can be derived on the importance of fault events in the proposed KG. Figure ~\ref{pr_alarmevents} shows the PageRank scores for various fault events and alarms in our KG. As can be visualised, events like \textit{Grid problems}, \textit{High temperature on the gearbox bearing} and \textit{Very high wind direction misalignment} are amongst the elements with highest importance, signifying the frequent nature as well as the complexity of O\&M activities for such cases. It is interesting to note that we observed that the identified fault events with higher PageRank scores have a greater number of interconnections with O\&M activity nodes (Preventive, Corrective and Predictive) and linkage to SCADA features, outlining that their influence in impacting outcomes from the KG are significant. O\&M engineers can use these insights to prioritise maintenance activities for the more critical fault events.

\subsection{Interfacing with Explainable AI models for decision support}
A primary advantage of KGs is that they are natively developed to facilitate explainability, with their entities and semantic technologies being a significant implementation of symbolic AI \cite{info11020122}. With the recent advent of XAI models in the wind industry \cite{BARREDOARRIETA202082}, achieving human-understandable decisions is key to facilitate trustworthy decision making for O\&M. The KG can serve as a powerful resource when used in combination with XAI models, as the important parameters and reasoning of the decisions made by the AI models can be used as the basis to retrieve appropriate maintenance actions directly from the KG. We discuss the integration of our KG with an XAI model for explainablity in autonomous decision support below. 
\paragraph{\textbf{XGBoost + SHAP for Explainable anomaly prediction}}
We propose the integration of our KG with XGBoost (eXtreme Gradient Boosting) \cite{10.1145/2939672.2939785}, an XAI model which has seen immense success in anomaly prediction with SCADA data in the wind industry in recent times \cite{8329419,app10093258,RePEc:gam:jeners:v:12:y:2019:i:22:p:4224-:d:283963}, including with data from the LDT \cite{windenergy_journal,renew_paper}. XGBoost utilises gradient boosting and ensemble learning to leverage the best predictions made by multiple decision tree models, and is highly computationally efficient and scalable. The XGBoost model can easily provide feature importances for the SCADA data to identify contributing factors which led to an anomaly in specific turbine sub-components, making it a viable and highly promising choice for explainable decision support. We also propose the utilisation of SHAP \cite{10.5555/3295222.3295230}, a specialised game-theory based approach which makes use of credit allocations and local explanations to explain the outputs of AI models.

We utilised the XGBoost + SHAP model proposed by Chatterjee and Dethlefs \cite{renew_paper} with 21,392 samples of SCADA observations from the LDT (at generally 10 minutes intervals), consisting of $102$ SCADA features and labelled history (ground truth) of faults across the $14$ different Functional Groups as discussed in Section ~\ref{data_sec}.
A training-test data split of $70-30\%$ was used, with $100$ individual decision trees as the learning estimators, a learning rate of $0.1$, $10$ early stopping rounds and a multi-class \textit{multi:softprob} objective function. The model achieves an accuracy of up to 99\% for anomaly prediction, with an F1 score of up to 0.99 \cite{renew_paper}. The model was developed in Python, and the code is made publicly available on GitHub \footnote{XAI4Wind Supplementary Resources: \url{http://github.com/joyjitchatterjee/XAI4Wind} \label{xai4wind_resources}}.

\paragraph{\textbf{Integration of XGBoost + SHAP model with the Knowledge Graph Database}}
Given that the XGBoost model was developed and trained in Python, whereas, the Neo4j graph database is conventionally a part of the Neo4j Desktop Application, we utilised \textit{Py2neo}\footnote{Py2neo Handbook: \url{https://py2neo.org/2020.1/} \label{py2neo_ref}}, a specialised Python library which facilitates connection of the Neo4j knowledge graph database server with Python applications. Based on the XGBoost model's predictions for the test dataset and the feature importances of top-10 SCADA parameters likely contributing to the fault, we created Cypher queries in Python to determine:-
\begin{enumerate}
    \item Preventive maintenance actions for the Functional Group in which the fault occurs.
    \item Corrective maintenance actions related to the Functional Group and fixing inconsistency in the SCADA features leading to the fault. 
    \item Predictive maintenance actions for Functional Groups through continuous monitoring of SCADA features for any inconsistencies which can lead to a fault.
\end{enumerate}
An example of a Cypher query for retrieving corrective maintenance actions from the KG and its execution in Python is shown in Listing ~\ref{correctivequery_eg}. Here, \textit{imp\_try} is a variable denoting the name of the feature which is likely leading to the anomaly, as identified by the XGBoost + SHAP model, and it is mapped to the SCADA features in the LDT dataset with the \$name parameter.

\begin{lstlisting}[float,belowskip=0pt,caption=Example Cypher query for extracting corrective actions from the KG in Python based on SCADA features,label=correctivequery_eg,basicstyle=\small,frame=single,breaklines=True]
  
query = "MATCH(n:Corrective)-[:ACTION]->(p)-[:FOR]->(q)-[:RELATESTO]-(r:Feature{name:$name}) RETURN p,q,r"
    
print(graph.run(query2, parameters= {"name": imp_try}).data())
\end{lstlisting}

Besides mapping the maintenance actions to the SCADA features, we also developed queries to automatically leverage the KG database in mapping predicted faults in the Functional Group to the appropriate O\&M actions. An example of this query is shown in Listing ~\ref{corr_fg}. Here, \textit{fno} is a variable which refers to the class of the Functional Group wherein the anomaly is predicted, and it is mapped to one of the 14 different Functional Groups in the LDT dataset using the \$fno parameter. When the graph query is excuted through \textit{graph.run}, the value of \textit{fno} passed is the predicted class in the test dataset (described as the value at \textit{y\_test} for the current sample under test for which the O\&M action is to be retrieved).

\begin{lstlisting}[float,belowskip=0pt,caption=Example Cypher query for extracting corrective actions from the KG in Python based on the Functional Group,label=corr_fg,basicstyle=\small,frame=single,breaklines=True]
  
query = "MATCH(n:Corrective)-[:ACTION]->(p)-[:FOR]->(q)-[:AFFECTS]-(r{fno:$fno}) RETURN p,q,r"

print(graph.run(query3, parameters= {"fno": int(y_test.iloc[current_sample_totest])}).data())

\end{lstlisting}

With the integration of the proposed KG with the XAI model, thorough descriptions of faults and appropriate maintenance actions can be generated at runtime (when the XAI model makes predictions with SCADA data). Below, we discuss 2 different cases of anomalies predicted by the XAI model, and examples of O\&M reports obtained:-

\begin{itemize}

\begin{figure}[!h]
\centerline{\includegraphics[width=1.2\textwidth,height=0.2\textheight]{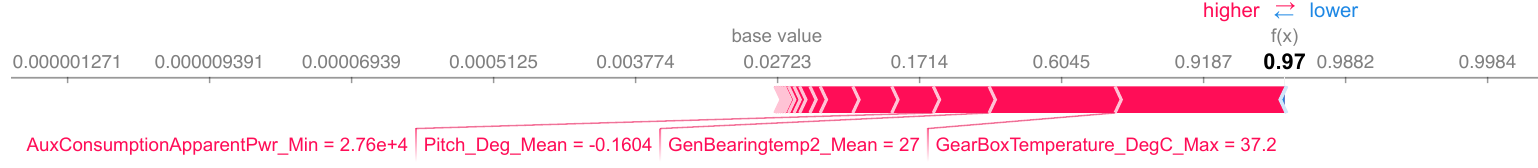}}
\caption{Force plot for the predicted gearbox anomaly, with the outlined features contributing positively (higher-red) to the fault \label{force_gearbox}}
\end{figure}

\item \textbf{Anomaly in Gearbox:} 
The first case we discuss is an anomaly predicted by the XAI model in the turbine's gearbox. Figure ~\ref{force_gearbox} shows the force plot outlining the contribution of SCADA features towards this prediction. The red color (higher metric) indicates that features such as the \textit{GearBoxTemperature\_DegC\_Max} contribute positively to this prediction, with a potential increase in the gearbox oil temperature causing the fault in the gearbox. There is a 97\% probability (0.97 as shown in the force plot) that these features would have a positive impact leading to a gearbox operational inconsistency. This is highly relevant to the gearbox anomaly based on existing literature, as there is a high degree of correlation between the gearbox oil and wind turbine gearbox conditions \cite{CORONADO2014747} and appropriate monitoring and management of such parameters can play a vital role in managing incipient gearbox failures \cite{en12224224,ZENG2020106233}. As an interesting point to note, our XAI model also indicates the change-point for the anomaly, including the exact values of the parameters at the time the anomaly occurs as can be seen from Figure ~\ref{force_gearbox}.

In Figure ~\ref{gearbox_pie}, the top-10 important features identified by the XGBoost+SHAP model for this prediction are described in terms of their percentage contribution (based on the feature importance values, generally referred to as the Shapley values \cite{10.5555/3295222.3295230}). Based on the top-10 important features, our integrated model automatically extracts the relevant nodes for the maintenance actions and the relevant fault event which is likely caused due to abnormality in the SCADA features. Listing ~\ref{gbox_nodeseg} describes examples of nodes automatically retrieved from the KG corresponding to the \textit{Gearbox Oil Sump Temperature Mean Value} SCADA feature which is identified to be a major contributor to the fault. It can clearly be seen that the most appropriate maintenance actions corresponding to the identified fault event (in this case \textit{High temperature on the gearbox oil}) are represented through the nodes. Note that these maintenance actions are specific to the fault type as well as the important SCADA feature, and provide the engineers \& technicians with guidance on how the operational inconsistency can be fixed/averted by handling the issues in the SCADA feature. It is interesting to note that besides the text-based descriptions of the maintenance actions, our model also provides links to the images which are relevant to the predicted anomaly - in this case redirecting the user to Gearbox Connection Diagrams and Gearbox Assembly as per Listing ~\ref{links_gearbox}. Figure ~\ref{gearassembly} shows an example visualisation for the Gearbox Assembly.

\begin{figure}[!h]
\centerline{\includegraphics[width=0.8\textwidth]{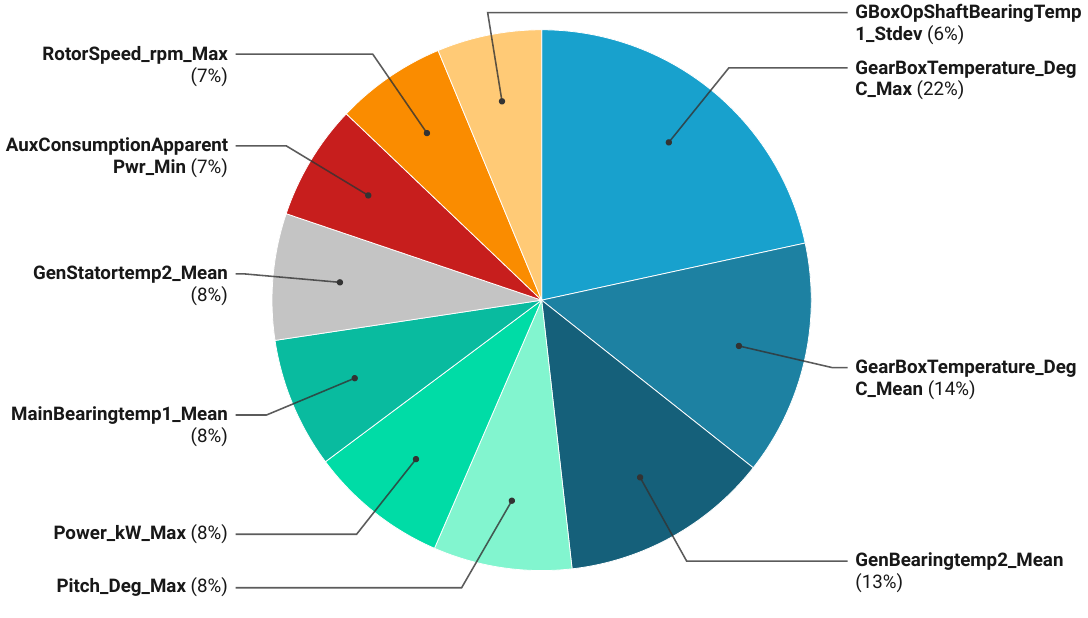}}
\caption{Pie chart outlining the percentage contribution of top-10 SCADA features to the predicted gearbox anomaly\label{gearbox_pie}}
\end{figure}

  \begin{lstlisting}[float,belowskip=0pt,caption=Example description of nodes extracted during gearbox anomaly,label=gbox_nodeseg,basicstyle=\small,frame=single,breaklines=True]
[{'p': Node('CorrAct25', activities=['Checking the adjustment of louvers', 'Gearbox PT sensor', .....]), 

'q': Node('FaultEvent25', details='High temperature on the gearbox oil'), 

'r': Node('Feature', description='Gearbox Oil Sump Temperature Mean Value', feature_no=51, name='GearBoxTemperature_DegC_Mean', unit='deg celsius')}, 

{'p': Node('CorrAct37', activities=['Resetting the switch/breaker', 'Checking the amp settings',....]),

'q': Node('FaultEvent37', details='Temperature switch of the gearbox pump')....]

\end{lstlisting}

 \begin{lstlisting}[float,belowskip=0pt,caption=Image reference property in the corrective maintenance action node for the gearbox anomaly,label=links_gearbox,basicstyle=\small,frame=single,breaklines=True]

image_url:['https://github.com/joyjitchatterjee/XAI4Wind/blob/master/images_maintenance/Gearbox_ConnectionDiagram.png',

'https://github.com/joyjitchatterjee/XAI4Wind/blob/master/images_maintenance/Gearbox_Assembly.png',

'https://github.com/joyjitchatterjee/XAI4Wind/blob/master/images_maintenance/Bearing_Change.png']
\end{lstlisting}
    
\begin{figure}[!h]
\centerline{\includegraphics[width=0.8\textwidth]{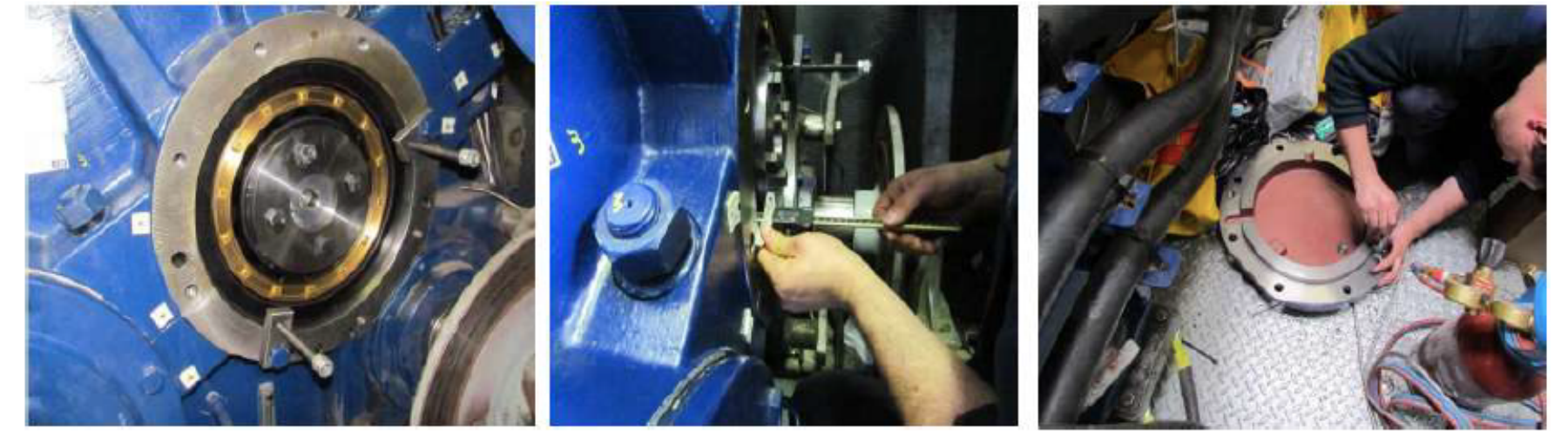}}
\caption{Visualisation of the Gearbox Assembly as obtained through the \textit{image\_url} attribute in the KG. The image is extracted from the Skillwind maintenance manual based on the predicted anomaly \label{gearassembly}}
\end{figure}    
    
As it is infeasible to show all corrective maintenance actions and the complete reports in this paper, we provide some examples of our model outputs for this operational inconsistency. Table ~\ref{gearboxcorr_eg1} describes some examples for different important features for the predicted gearbox anomaly, the corresponding fault events which are relevant and a subset of the appropriate maintenance actions.

\begin{table}[!h]
\tiny
\caption{Examples of corrective maintenance actions based on identified important SCADA features for the predicted gearbox anomaly\label{gearboxcorr_eg1}}
    \begin{tabular}{ p{4cm}  p{3.4cm}  p{5cm} }
        \toprule
\textbf{Important Feature}     
&  \textbf{Relevant Fault Event} 
&  \textbf{Retrieved Examples of Corrective Action(s)} \\\midrule
GearBoxTemperature\_DegC\_Max (Gearbox Oil Sump Temperature Maximum Value)
&  High temperature on the gearbox oil & 1. Checking the adjustment of louvers. 2. Checking the wires for damages and appropriate connections. 3. Making sure the bearing temperature is higher than oil temperature, if not the wiring could be swapped. 4. Making sure the PLC is reading oil temperature correctly. If it is too high, it implies that the pump is running too often. \\\hline
GenBearingtemp2\_Mean (Generator Bearing 2 Temperature Mean Value)
& High temperature on bearings (DE or NDE) & 1. Listening for any unusual noise coming from bearing (could be a bad bearing). 2. Checking the temperature touchscreen. Replacing the PT100 or temperature card. 3. Making sure the lubber has enough grease and it is flowing into the bearing.  \\\hline
GenStatortemp2\_Mean (Generator Stator Temperature Mean Value)
& IGBT high temperatures & 1. High temperature on INU IGBT's. 2. Checking and saving the fault log and data logger.  3. Checking the ambient temperatures.4. Checking the air flow and fan operation. 5. Checking and clearing air filters in the cabinet and drivers   \\\hline
GBoxOpShaftBearingTemp1\_Stdev (Gearbox Bearing 1 Temperature Standard Deviation)
& High temperature on the gearbox bearing & 1. Looking for bearing damages. 2. Checking the multiplier pump/ cooling units. 3. Checking the wiring and cables for damages.\\
        \bottomrule
    \end{tabular}
\end{table}

A specialised provision which our model provides is the additional ability to generate maintenance actions based on the features which may not directly contribute to the fault in an obvious manner, but are identified as important features by the XAI model. For instance, \textit{Pitch\_Deg\_Max} (denoting the pitch angle maximum value) is identified to have a contribution of 8\% towards the prediction by our model, however, there is no direct relationship between the pitch angle with a gearbox anomaly based on general domain understanding. This feature can indeed have indirect relevance to the operational inconsistency, as in some instances like wind turbine blade icing, the aerodynamic performance of the blades can be affected leading to operational inconsistencies including power loss, equipment failure, mechanical failure etc \cite{en11102548}. Further, the mass and aerodynamic imbalance potentially leads to loading in the gearbox, which can cause the predicted anomaly \cite{en9110862}. In such cases, maintenance actions specific to the identified (non-obvious) features can be helpful for the engineers \& technicians to fix any inconsistency which may have a hidden association with the predicted anomaly. Also, since AI models are not perfect in making predictions and false alarms can arise, identifying maintenance actions specific to important features (rather than being only specific to the predicted anomaly) can help O\&M engineers make critical decisions in such circumstances. As an example, maintenance actions for the fault event \textit{Possible existence of ice on blades} is shown in Listing ~\ref{pitchgear_lst}.

\begin{lstlisting}[float,belowskip=0pt,caption=Example of corrective maintenance actions for Pitch Angle Maximum Value SCADA feature potentially occurring due to ice existence on blades and leading to an indirect anomaly in the gearbox,label=pitchgear_lst,basicstyle=\small,frame=single,breaklines=True]

{'p': Node('CorrAct44', activities=['Checking the rotor sensor connection', 'Checking for loose connections in the A-9 box', 'Checking for damaged cables', 'Verifying the wiring', 'Visually inspecting the blades for ice build up'])}
\end{lstlisting}

Besides the corrective maintenance actions, our proposed methodology is also able to extract relevant predictive and preventive actions for the identified anomaly in the gearbox. Few examples of preventive maintenance actions for averting operational inconsistency in the gearbox extracted by our model are shown in Table ~\ref{prev_gearbox_eg1}. Alongside the brief description for these actions, our model also provides the general and initial periodicity (wherever applicable as per the Skillwind manual) to help engineers make appropriate and timely decisions.

\begin{table}[h]
\caption{Examples of preventive maintenance actions extracted by the XAI-KG integrated model for the gearbox\label{prev_gearbox_eg1}}
\resizebox{\textwidth}{!}{%
\begin{tabular}{|l|l|}
\hline
\textbf{Preventive Activity Details} & \textbf{General Periodicity} \\ \hline
Gearbox oil replacement & 48 months \\ \hline
Gearbox vent filter replacement & 12 months \\ \hline
Gearbox supports inspection & 12 months \\ \hline
Cooling system for gearbox- General inspection & 12 months \\ \hline
Cooling system for gearbox- Oil filter substitution & 12 months \\ \hline
\end{tabular}%
}
\end{table}

\begin{figure}[!h]
\centerline{\includegraphics[width=1.2\textwidth,height=0.2\textheight]{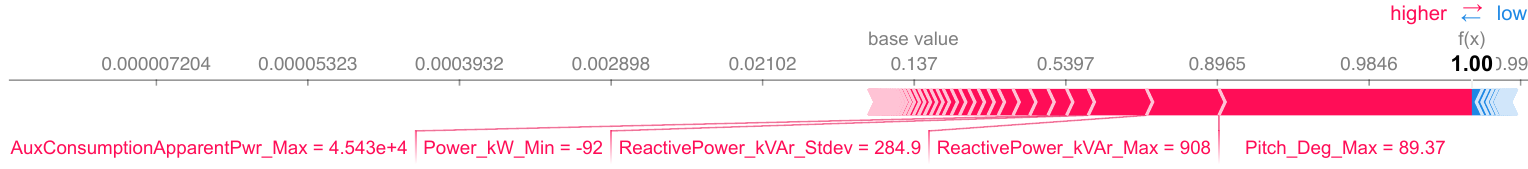}}
\caption{Force plot for the predicted pitch system anomaly, with the outlined features contributing positively (higher-red) or negatively (low-blue) to the fault \label{force_pitch}}
\end{figure}

\item \textbf{Anomaly in Pitch System:} The second case is for an anomaly predicted by the XAI model in the \textit{Pitch System Interface Alarms} functional Group. Similar to the discussion above for the gearbox, refer Figure ~\ref{force_pitch} for the force plot outlining the distribution of SCADA features leading to the pitch system anomaly. Amongst notable features, \textit{Pitch\_Deg\_Max} (denoting the pitch angle maximum value) shares a significant positive contribution (higher-red) to the fault, but can also have a negative impact (as the pitch angle generally has to be within an optimum range for normal operation of the turbine). The outlined features in the force plot show that the prediction has a probability of 1 (sure event), and this can help engineers \& technicians to be especially prepared for the anomaly, and thereby plan O\&M actions in advance to fix/avert the fault. Similar to Case 1 discussed above, the change-point for the features e.g. the pitch angle and its exact value during the prediction can also be clearly visualised.

The percentage composition of the top-10 contributing important SCADA features which lead to the prediction as identified by the XGBoost + SHAP model are shown in Figure ~\ref{pitch_pie}. As can be seen, \textit{Pitch\_Deg\_Max} accounts for 40\% of importance in the prediction, which makes it a vital parameter to monitor and fix any abnormality during O\&M. There are other features like \textit{WindSpeed\_mps\_Mean}, \textit{Power\_kW\_Stdev} and \textit{ReactivePower\_kVAr\_Max} which are comparatively less relevant but still important, as based on existing literature, wind speed, output power and grid-connective frequency (and thereby reactive power) are highly significant during the pitch system failure and reflect change in the system's operational behaviour \cite{en12142693}. 

\begin{figure}[!h]
\centerline{\includegraphics[width=0.8\textwidth]{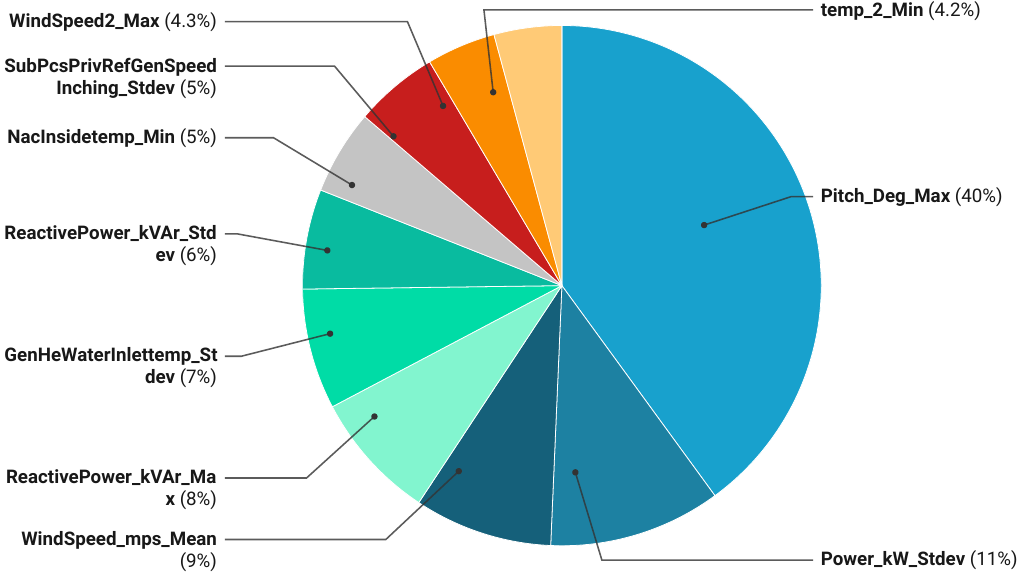}}
\caption{Pie chart outlining the percentage contribution of top-10 SCADA features to the predicted pitch system anomaly\label{pitch_pie}}
\end{figure}

% An example of a general corrective maintenance action for the pitch system (without considering the SCADA features) is shown in Listing ~\ref{}. This specialised ability of our model can be extremely useful in situations wherein the XAI model is not able to identify the important features correctly (as mentioned before, AI models are not perfect), and can help support O\&M decisions specific to the sub-component rather than simply judging the situation based on the SCADA features.  

By considering the top-10 important SCADA features, similar to Case 1 discussed above, our model is able to generate informative descriptions of maintenance actions towards fixing/avert the pitch system anomaly. See Listing ~\ref{pitchcorreg_lst} for an example of generated corrective actions for an  abnormality in the \textit{Pitch\_Deg\_Max} feature, which is potentially contributing to the \textit{Blade Position Error} fault event in the pitch system.

\begin{lstlisting}[float,belowskip=0pt,caption=Example description of nodes extracted during pitch system anomaly caused due to Blade Position Error,label=pitchcorreg_lst,basicstyle=\small,frame=single,breaklines=True]
{'p': Node('CorrAct45', activities=['Checking the power supply', 'Checking the cables and electrovalves for damage', 'Verifying the I/O modules in the hub', Swapping the cards between blades to see if fault is associated to one specific blade.', 'Verifying the balluf settings']), 

'q': Node('FaultEvent45', details='Blade Position Error'), 

'r': Node('Feature', description='Pitch Angle Maximum Value', feature_no=2, name='Pitch_Deg_Max', unit='Deg')}
\end{lstlisting}

Examples of various types of corrective maintenance actions for different fault events and identified important features for the predicted pitch system anomaly are shown in Table ~\ref{pitchcorr_eg2}. It is integral to note, that in a few cases (e.g. for the feature \textit{Power\_kW\_Stdev} denoting the \textit{Active power standard deviation} in this case), it is possible that a feature may not have an associated maintenance action in the KG database wherein the feature does not have a direct attribute for maintenance activities and is dependent on other SCADA features (e.g. the turbine active power depends on features like wind speed and pitch angle), and any corrective action for the contributing features would directly fix/avert any inconsistency in the determined important feature.  In such cases, our model extracts the next highest-priority feature based on its importance score (Shapley value) to provide the appropriate maintenance strategies.

\begin{table}[!h]
\tiny
\caption{Examples of corrective maintenance actions based on identified important SCADA features for the predicted pitch system anomaly\label{pitchcorr_eg2}}
    \begin{tabular}{ p{4cm}  p{4cm}  p{5cm} }
        \toprule
\textbf{Important Feature}     
&  \textbf{Relevant Fault Event} 
&  \textbf{Retrieved Examples of Corrective Action(s)} \\\midrule
Pitch\_Deg\_Max\\ (Pitch Angle Maximum Value)
& Pitch Activation Error. \tablefootnote{If the pitch of any of the blades differs some degrees from the reference pitch for a short period.}     
& 1. Making sure the balluf cable is not disconnected.  2. Checking the balluf condition. 3. Checking proportional valve or cable. 4. Checking the relays for each blade in the hub cabinet. \\\hline
WindSpeed\_mps\_Mean (Average Wind Speed Mean Value)
& High wind speed. \tablefootnote{This alarm is activated if the wind speed is above the design levels. }& 1. Checking the on-site wind speed. 2. Checking the wires and voltages in the connection box. 3. Checking for ice build up on the Sonic and Anemometer.  \\\hline
GenHeWaterInlettemp\_Stdev (Generator Inlet Temperature Standard Deviation)
& High temperature on generator’s slipring  & 1. Opening the louvers on generator. 2. Checking the motor fan condition.  3. Checking the PT sensor. 4. Checking the Temperature I/O card.  \\\hline
NacInsidetemp\_Min (Nacelle Temperature Minimum Value)
& High or Low nacelle temperature  & 1. Checking the temperatures.  2. Checking the PT sensor wiring. 3. Checking the PT placement. 4. Checking the PT card connections.
\\
        \bottomrule
    \end{tabular}
\end{table}

Besides the corrective maintenance actions, we provide examples of preventive maintenance actions generated by our integrated model in Table ~\ref{prev_pitch_eg2}. These are specific to the pitch system, and can serve as an integral resource for the O\&M engineers to prepare in advance for any impending faults, through appropriate monitoring and repair/fix activities wherever necessary.

\begin{table}[h]
\caption{Examples of preventive maintenance actions extracted by the XAI-KG integrated model for the pitch system anomaly\label{prev_pitch_eg2}}
\resizebox{\textwidth}{!}{%
\begin{tabular}{|l|l|}
\hline
\textbf{Preventive Activity Details} & \textbf{General Periodicity} \\ \hline
Pitch re-tightening & 12 months \\ \hline
Pitch calibration & 12 months \\ \hline
Batteries substitution & 48 months \\ \hline
Inspection of the clearance between pinion and crown & 12 months \\ \hline
Gear oil substitution
 & 60 months \\ \hline

\end{tabular}%
}
\end{table}

\end{itemize}

\section{Discussion}\label{discussion}
Through the use-cases discussed above, we have demonstrated the role which the proposed KG can play for supporting O\&M in the wind industry. We have also shown that the novel insights and metrics derived from the KG are highly relevant and informative for decision support in the wind industry based on existing literature. Also, as the maintenance actions are retrieved from a domain-specific resource in the wind industry (the Skillwind manual), the derived actions when used in conjunction with a real-world operational turbine data (from the LDT) can enhance the capabilities of the KG even further for autonomous decision support, beyond simply information querying and retrieval when used standalone. 

Note that while we have demonstrated the use-cases with SCADA data from the LDT, the proposed KG is not restricted to any specific wind turbine. The KG can serve as a generic resource for O\&M in the wind industry, and it can be integrated with SCADA data from any wind turbine to provide maintenance actions specific to predicted faults in the turbine. Similarly, the capabilities of the XAI model are not limited to the LDT data, but can easily be transferred to new turbines with their specific datasets. Also, the proposed KG can be potentially integrated with XAI models even in cases with absence of available data, through specialised techniques like transfer learning which has seen promising applications in the wind industry \cite{CHEN20212053,windenergy_journal}. 

\section{Conclusion}\label{conclusion}
We have showcased the immensely powerful role which multimodal KGs can play for explainable decision support in the wind industry. To the best of our knowledge, this is the first paper in the wind energy sector to propose real-world applications of KGs in O\&M of wind turbines, demonstrated through several real-world use-cases. By combining heterogeneous data in the wind industry, such as SCADA parameters with natural language based maintenance actions and images, we have shown the ability of the KG to be queried interactively as well as automatically through its integration with an XAI model. The proposed KG is able to provide effective human-intelligible maintenance action strategies based on various faults in turbine sub-components, which can be better analysed and interpreted by leveraging graph data science algorithms and XAI models which we have discussed through promising experimental observations in the paper. We envisage that the proposed KG, which is made publicly available, can serve as a driving resource to encourage future research in this area, helping  pave way for autonomous explainable decision support for CBM in the wind industry. 


\section*{Declaration of competing interest}
The authors declare that they have no known competing financial interests or personal relationships that could have appeared to influence the work reported in this paper.

\section*{Funding source}
This research did not receive any specific grant from funding agencies in the public, commercial, or not-for-profit sectors.

\section*{Data Availability}
The knowledge graph proposed in this paper is publicly available, and can be found on GitHub at\\ http://github.com/joyjitchatterjee/XAI4Wind.

\section*{Acknowledgement}
We would like to acknowledge the Skillwind project \footnote{Skillwind project: \url{https://skillwind.com} \label{skillwind_project}} for the publicly available Skillwind Maintenance Manual, which was integral to develop the maintenance actions segment of the KG. We are also grateful to ORE Catapult for providing us data from the Levenmouth Demonstration Turbine \footnote{ORE Catapult Platform for Operational Data: \url{https://pod.ore.catapult.org.uk} \label{pod_catapult}}, which helped us conduct experiments to demonstrate applications of the KG in real-world for the wind industry.

\section*{CRediT authorship contribution statement}
\textbf{Joyjit Chatterjee:} Conceptualization, Methodology, Software, Formal analysis, Investigation, Writing - Original Draft. \textbf{Nina Dethlefs:} Conceptualization, Supervision, Writing - Review \& Editing.

\bibliographystyle{elsarticle-num} 
\bibliography{main}

\appendix
\pagebreak

\section{Details of Nodes in the Knowledge Graph}
\label{kg_nodes_details}

% \usepackage{graphicx}
\begin{longtable}{|p{3.4cm}|p{4cm}|p{7cm}|}
\hline
\textbf{Node label} & \textbf{Description} & \textbf{Properties} \\ \hline
System & Reference to the Study Turbine (LDT) & location, name, rated\_power, type \\ \hline
Environment & Reference to the Study Turbine Environment & name \\ \hline
Blades & Wind Turbine Blades Sub-System & name, InspectionActivities, CorrectiveActivities, PreventiveActivities, \newline visualinspection\_image\_url,image\_url \\ \hline
ESC & Electric, Sensor \& Control Sub-System & name \\ \hline
FCS & Foundation \& Concrete Section Sub-System & \begin{tabular}[c]{@{}l@{}}name, image\_url, \\ CorrectiveActivities\end{tabular} \\ \hline
HydraulicSystem & Hydraulic Sub-System & \begin{tabular}[c]{@{}l@{}}name, image\_url, \\ CorrectiveActivities\end{tabular} \\ \hline
CommNetwork & Communications \& Network Sub-System & \begin{tabular}[c]{@{}l@{}}name, image\_url,\\ CorrectiveActivities\end{tabular} \\ \hline
Converter & Converter Sub-System & \begin{tabular}[c]{@{}l@{}}name, PredictiveActivities, \\ PreventiveActivities\end{tabular} \\ \hline
DriveTrain & Drive Train Sub-System & name, PreventiveActivities \\ \hline
Yaw & Yaw Sub-System & name, fno \\ \hline
ParkBrake & Park Brake Sub-System & name \\ \hline
YawBrake & Yaw Brake Sub-System & name, fno \\ \hline
MainShaft & Main Shaft Sub-System & name \\ \hline
PowerCabinet & Power Cabinet Sub-System & name \\ \hline
PitchSystem & Pitch Sub-System & name \\ \hline
Transformer & Transformer Sub-System & \begin{tabular}[c]{@{}l@{}}name, image\_url,\\ CorrectiveActivities\end{tabular} \\ \hline
FunctionalGroup & Reference to all Functional Groups & name, contents \\ \hline
IPR & IPR Functional Group & name, fno \\ \hline
PitchInterfaceAlarm & Pitch System Interface Alarms Functional Group & name, fno \\ \hline
PitchEFCMon & Pitch System EFC Monitoring Sub-System & name, fno \\ \hline
PCS & Power Conditioning System Functional Group & name, fno \\ \hline
PPD & Partial Performance Degraded Functional Group & name, fno \\ \hline
Yaw & Yaw System Functional Group & name, fno \\ \hline
HydraulicSys & Hydraulic System Functional Group & name, fno \\ \hline
Gearbox & (i) Gearbox Functional Group (ii) Gearbox Sub-System & \begin{tabular}[c]{@{}l@{}}(i)name, fno (ii)name, image\_url,\\ fno, CorrectiveActivities\end{tabular} \\ \hline
NoFault & No Fault Functional Group & name, fno \\ \hline
WindAlr & Wind Condition Alarms Functional Group & name, fno \\ \hline
GLLP & Ground Line \& Lightning Protection Sub-System & name \\ \hline
MVTR & Moisture Vapour Transmission Rate Functional Group & name, fno \\ \hline
Test & Test Rig Functional Group & name, fno \\ \hline
Pitch & Pitch System Functional Group & name, fno \\ \hline
SCADA & Reference to all SCADA features & name \\ \hline
Alarm1 to Alarm26 (both inclusive) & 26 different alarm types in the LDT & description, alarm\_no \\ \hline
Feature & Reference to distinct SCADA features & name, description, unit, feature\_no \\ \hline
MaintenanceAction & Reference to all maintenance actions & name, contents \\ \hline
FaultEvents & Reference to all fault events & name \\ \hline
FaultEvent1 to FaultEvent57 (both inclusive) & 57 different types of fault events & details \\ \hline
Preventive & Reference to all preventive actions & \begin{tabular}[c]{@{}l@{}}name, Lineups,\\ DriveTrainActivities, Cleaning,\\ CheckWearSlackLineups, \\ImpParameters,OilChanges,\\ SettingPressures, FunctionalChecks,\\ Filters, Corrosion, LostScuffsOrGaps, \\FatLiqours, Sampling, Retightening,\\ PlasticsOrDegradedGums\end{tabular} \\ \hline
PrevAct1 to PrevAct233 (both inclusive) & 233 distinct preventive maintenance actions & \begin{tabular}[c]{@{}l@{}}name, gen\_periodicity, details,\\ act, initial\_periodicity,\\ activities\end{tabular} \\ \hline
Predictive & Reference to all predictive actions & name, contents \\ \hline
PredAct1 to PredAct11 (both inclusive) & 11 distinct predictive maintenance actions & details, activities, image\_url \\ \hline
Corrective & Reference to all corrective actions & name, contents \\ \hline
CorrAct1 to CorrAct57 (both inclusive) & 57 distinct corrective maintenance actions & activities, image\_url \\ \hline
\end{longtable}





% \end{thebibliography}
\end{document}